\documentclass[runningheads]{llncs}
\usepackage[T1]{fontenc}
\usepackage{graphicx}
\usepackage{amsmath}
\usepackage{amssymb}
\usepackage{wrapfig}
\usepackage{placeins} 

\author{Aniruddha Maiti\inst{1}\orcidID{0000-0002-1142-6344} \and
 Satya Nimmagadda\inst{2}  \and
 Kartha Veerya Jammuladinne\inst{1} \and
  Niladri Sengupta\inst{3}  \and 
 Ananya Jana\inst{2}}  
\institute{
West Virginia State University, Institute, WV 25112\\
\email{\{aniruddha.maiti, kjammuladinne\}@wvstateu.edu}
\and
Marshall University, Huntington, WV\\
\email{\{jana, nimmagadda2\}@marshall.edu}
\and
Fractal Analytics Inc., USA\\
\email{dinophysicsiitb@gmail.com}
}

\authorrunning{Maiti et al.}

\begin{document}
\title{Convergence of Outputs When Two Large Language Models Interact in a Multi-Agentic Setup}

\titlerunning{Convergence of LLMs in Multi-Agentic Setup}
\maketitle
\begin{abstract}
In this work, we report what happens when two large language models respond to each other for many turns without any outside input in a multi-agent setup. The setup begins with a short seed sentence. After that, each model reads the other's output and generates a response. This continues for a fixed number of steps. We used \textbf{Mistral Nemo Base 2407}  and \textbf{Llama 2 13B hf}. We observed that most conversations start coherently but later fall into repetition. In many runs, a short phrase appears and repeats across turns. Once repetition begins, both models tend to produce similar output rather than introducing a new direction in the conversation. This leads to a loop where the same or similar text is produced repeatedly. We describe this behavior as a form of convergence. It occurs even though the models are large, trained separately, and not given any prompt instructions. To study this behavior, we apply lexical and embedding-based metrics to measure how far the conversation drifts from the initial seed and how similar the outputs of the two models becomes as the conversation progresses.

\textbf{Keywords:} Convergence in Multi-Agent, Agentic Conversation, Multi-Agent, Multi-LLM Interaction.
\end{abstract}

\section{Introduction}

Most evaluations of large language models rely on short prompts or single-turn completions. These tests measure the correctness, fluency, or instruction following capability in isolation. They do not reveal what happens when a model must respond to its own output or engage in a long exchange with other models. Prior works indicate that dialogue diversity tends to degrade over long-term simulations \cite{chu2024exploring}. This suggests that the behavior of language models over time may expose failure modes that are not visible in one-step settings.

This work extends that idea to a two-model setup. Instead of using a single model recursively, we allow two different models to take turns responding to each other. Each model runs in its own process, with separate weights and tokenizers, and reads only the plain-text output of the other. There is no shared memory, no prompts, and no injected system instructions. The setup is minimal: the models are connected only through raw text files.

The models used are \textit{Mistral Nemo Base 2407} and \textit{Llama 2 13B hf}. The original Mistral 12B parameter model is developed by Mistral AI and NVIDIA. Different users have developed a variety of fine-tuned versions of this base model for different use-cases since its publication. The original Llama-2 13 Billion Parameter model is released by Meta. Similar to Mistral, different users and developers have fine-tuned or adopted the original base model for a variety of purposes. These two models are large autoregressive transformers with different training sources and configurations. Their architectural similarity and difference in training data during training make them a useful pair for studying how agent-like conversation progresses when they respond alternately based on the other model's output.

This setting raises a natural question: when two models interact only through language, how long can the conversation remain meaningful? Will they maintain topic and coherence, or will they collapse into repetition? If so, when and how does that collapse occur?

While exploring these questions, we found that model pairs begin with coherent dialogue, but often drift into repetition after a few turns. In some cases, the collapse is gradual. In others, it is sudden and marked by a repeated phrase. Once repetition sets in, it tends to persist in the following turns. This behavior appears even though the models are large, capable, and not aligned through any shared context or fine-tuning.

Studying this interaction gives insight into the stability and limits of generative systems. It also helps identify convergence behaviors that are easy to miss in prompt-based evaluations. Understanding how and when models fall into low-diversity states is important for tasks that involve long-form generation or multi-agent communication. 

\section{Related Work}

Research on multi-agent large language models (LLMs) has expanded rapidly in the past two years. This is a recent trend that investigates how several LLMs interact. Earlier studies differed in goals, but most of them examined how multiple models exchanged messages, reasoned together, or produced stable output. The main lines of work are in the areas of : frameworks for multi-agent interaction, evaluation methods, dialogue stability, and conversational control.

Some studies proposed frameworks for linking multiple LLMs. AutoGen \cite{wu2024autogen} described a setup where agents exchanged both code and natural language messages. The framework handled structured communication and role assignment. Song et al. \cite{song2024adaptive} presented a related design involving a primary agent that forms and manages smaller teams during a task. Guo et al. \cite{guo2024large} provided a survey that summarized agent structures and their communication strategies. Li et al. \cite{li2025advancing} developed a reinforcement-learning method that assigned and refined roles automatically through training. Yuan and Xie \cite{yuan2025reinforce} modeled a feedback process where agents improved reasoning over multiple steps. White et al. \cite{white2025collaborating} applied such multi-agent systems to an embodied task in Minecraft.

Evaluation of multi-agent conversations is another active area of research. Guan et al. \cite{guan2025evaluating} and Chan et al. \cite{chan2023chateval} did work in this direction. Chuang et al. \cite{chuangdebate} released a benchmark for simulated long-form discussions.  Ku et al. \cite{ku2025multi} investigated debate between two agents. 

Some studies examined long-term behavior and repetition. Chu et al. \cite{chu2024exploring} studied diversity loss in agent dialogue. Chu et al. \cite{chu2024cohesive} investigated repetition and factual errors in simulated conversations. Becker \cite{becker2024multi} analyzed how agents performed on different tasks. Maharana et al. \cite{maharana2024evaluating} evaluated memory over hundreds of conversational turns. Han et al. \cite{han2024llm} discussed open challenges in managing shared context and reasoning across multiple agents.

Some studies addressed control and consistency in agentic setup. Darwish et al. \cite{darwish2025mitigating} proposed a rule-based setup that constrained agent outputs and improved reliability. Guo et al. \cite{guo2024large} and Han et al. \cite{han2024llm} both discussed safety risks in open-ended multi-agent dialogue. Becker \cite{becker2024multi} reported that shorter conversations reduced these failures.

Most prior work examined collaboration, evaluation, or coordination under defined tasks. Very few studies analyzed what happens when two independent models converse freely without supervision. The present work investigated this open setting by examining what happens when two LLMs interacted for many turns. 

\section{Experimental Setup}
\begin{figure}[htbp]
    \centering
    \includegraphics[width=0.9\textwidth]{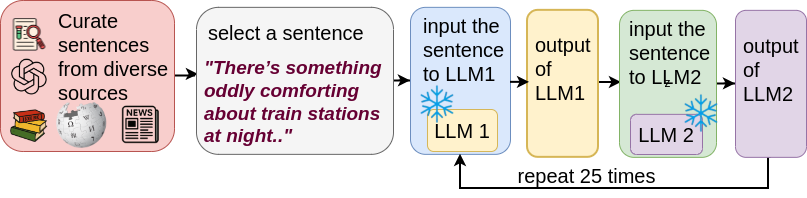}
    \caption{Experimental pipeline showing how two large language models interact over multiple turns. Each model alternately reads the other’s output and generates a new response, continuing until the defined number of steps is reached.}
    \label{fig:pipeline5}
\end{figure}

To examine how two large language models behave when they respond to each other over multiple turns without any external input or correction, we created the following experimental setup. We took two models which were pre-trained independently. We make them run in their own distributed processes. The conversation begins with a short seed sentence. After that, the models take turn reading and replying to each other’s outputs for twenty-five times. No prompts, templates, or system instructions were added. The input to each model is the raw text produced by the other. The pipeline of the experimental setup is depicted in Fig.\ref{fig:pipeline5}

The models used in this study are Model A, \textbf{Mistral Nemo Base 2407} (12 billion parameters, developed by Mistral AI and NVIDIA) and Model B, \textbf{Llama 2 13B hf} (13 billion parameters, released by Meta) \cite{touvron2023llama}. Both are transformer-based language models. They are open-weight and widely used. Although the architectures are similar, they differ in training data, optimization choices, and tokenizer vocabulary. These differences make them suitable for studying how independently trained systems interact through natural language alone.

We used HuggingFace’s \texttt{transformers} library to load the models. We also leveraged PyTorch’s\textbf{ Fully Sharded Data Parallel (FSDP) framework}, and mixed precision. Model parameters used \texttt{bfloat16}, and reductions are computed in \texttt{float32}. Models are wrapped using a mixed-precision policy and sharded across two GPUs per model \cite{zhao2023pytorch}. Mistral runs on devices 0 and 1, while Llama runs on devices 2 and 3. The two processes do not share memory or internal states. They only communicate through file I/O using plain UTF-8 text files. 

Each experiment begins by selecting a seed sentence from a predefined list. This sentence is written to disk as the starting input. Model A reads the file, generates a continuation using the fixed generation parameters, and writes its output to a new file. Model B then reads Model A’s output and generates its own response. This back-and-forth process continues for 25 turns, alternating between models.

To coordinate the sequence, file existence is used as a signal. Each process waits for the input file to appear before continuing. After reading the input and generating a response, each model writes its output and waits at a synchronization barrier. This ensures that outputs are ordered and deterministic and there is no overlap in the execution of the models.

We used nucleus sampling with \texttt{top-p} = 0.95 and \texttt{temperature} = 0.7 \cite{wang2025msarl}. Each generation is capped at maximum 50 new tokens. 

All outputs (after removing the input string from the output) are saved as-is, without filtering, cleaning, or normalization. Each round is stored in its own directory. This isolated setup captures only the natural text-level interaction between the two models.

For consistency, we define the following terminologies: A \textbf{turn} is defined as two consecutive steps: the first one by \emph{Mistral Nemo Base 2407} (Model A) and the second one by \emph{Llama 2 13B hf} (Model B). Starting from each seed sentence, we completed 25 such turns. Each such 25  turns is collectively defined as a \textbf{round} representing a full conversation. A \textbf{step} is defined as one inference step of either of the two models. Therefore, each \textbf{round} is consisted of $25$ \textbf{turns} or $50$ \textbf{steps}.  

\subsection{Origin or Seed Sentences}

Each round begins with a single seed sentence. This sentence is used as the initial input to Model A at turn 1. We use a total of 50 different seeds in 50 different rounds. The seeds come from five sources. Each source contributed 10 sentences. Table~\ref{tab:seed_sources} lists the sources and describes their content.

\begin{table}[h]
\centering
\begin{tabular}{|l|p{10cm}|}
\hline
\textbf{Source} & \textbf{Description} \\
\hline
\textit{Prompt generated} & Sentences generated using the GPT-4 Omni Web-based API. The used prompt describes the two-model set-up briefly and ask to generate seed sentence for such set-up. \\
\hline
\textit{Wikipedia} & First sentences from different Wikipedia entries. These cover science, history, and popular topics. \\
\hline
\textit{News articles} & First sentences from recent news articles. Topics include markets, climate, and global affairs. Sentences are factual and time-specific. \\
\hline
\textit{Scientific papers} & Sentences from scientific abstracts and introductions. These are often longer and technical. Chosen sentences cover scientific fields such as physics, biology, and computer science. \\
\hline
\textit{Novels} & First sentences from published fiction. These include a range of styles and time periods. Some are simple. Others are syntactically complex. \\

\hline
\end{tabular}
\caption{Seed sentence sources. We have taken 10 seed sentences from each source.}
\label{tab:seed_sources}
\end{table}

We selected seeds that differ in style, structure, and subject. Some are short, the others contain discipline-specific terms or longer clauses. This variety helps examine whether the form of the seed changes the behavior of conversation dynamics. It also helps detect whether convergence occurs more often with certain sentence types. 

\section{Analysis and Metrics}
\label{sec:metrics}
To understand how the conversation generated by two models evolve, we analyze the generated sequences using multiple quantitative metrics. The quantification is done in different spaces such as lexical space, space of strings or tokens, embedding space etc. Four metrics have been used for this purpose : Cosine distance, Jaccard distance, distance based on BLEU score \cite{papineni2002bleu}, difference in Coherence. We describe each of these metrics briefly below.

\subsection{Cosine Distance in Embedding Space}
To measure semantic drift and convergence, we use sentence embeddings and cosine distance. Let $s \in \mathbb{R}^d$ denote a sentence embedding in a $d$-dimensional vector space. For two embeddings $s_1$ and $s_2$, their cosine distance is defined as:
\[
D_{\text{cos}}(s_1, s_2) = 1 - \frac{s_1 \cdot s_2}{\|s_1\| \, \|s_2\|}.
\]
This distance reflects the difference between two texts in terms of their semantic content, independent of their exact lexical form. Cosine distance in embedding space is useful because convergence may be semantic rather than lexical. Embedding-based comparisons detect such cases. The sentence embeddings have been calculated using \emph{all-mpnet-base-v2} \cite{siino2024all} embedding model.

\subsection{Jaccard Distance Over Tokens}
We computed the Jaccard distance between token sets from each model's output in two consecutive steps \cite{singh2021text}. For two token sets \( A \) and \( B \), the Jaccard similarity is defined as:
\[
J(A, B) = \frac{|A \cap B|}{|A \cup B|}.
\]
This value ranges from 0 (no overlap) to 1 (identical sets). We used $1-J(A, B) $ as a distance measure. It provides an additional view of how much consecutive output strings are different from each other.

\subsection{BLEU Overlap Across Turns}
We used BLEU \cite{papineni2002bleu} scores between consecutive model outputs. Given reference text \( R \) and candidate text \( C \), the BLEU score is a weighted geometric mean of modified \( n \)-gram precisions, penalized by a brevity factor:
\[
\text{BLEU}(C, R) = \text{BP} \cdot \exp \left( \sum_{n=1}^{N} w_n \log p_n \right),
\]

where \( p_n \) is the modified precision for \( n \)-grams of order \( n \), and \( w_n \) is the weight assigned to each order (usually uniform). BLEU helps identify overlap between short phrases. High BLEU scores indicate convergence to a fixed pattern. We used $1- \text{BLEU}(C, R)$ as a distance metric to quantify the difference between two consecutive outputs.

\subsection{Difference in Coherence}
We also used the change in coherence as a metric. For each step \( t \), we computed a coherence score between two consecutive outputs. We used a bigram-based normalized pointwise mutual information (NPMI) approximation to estimate this coherence \cite{bouma2009normalized}. This provides a single scalar score that reflects how likely the sentence transitions are within a local context.

Let \( C(x_t, y_t) \) denote the coherence score between input and response at step \( t \). We measured the absolute change in coherence between consecutive steps:
\[
\Delta_{\text{coh}}(t) = |C(x_{t+1}, y_{t+1}) - C(x_t, y_t)|.
\]
This score tracks whether the sentence-to-sentence relationship is becoming more or less coherent over time.

No single metric captures all aspects of convergence. Lexical measures (BLEU, entropy, Jaccard) are sensitive to repetition of the same words or tokens but can miss semantically similar rewording or paraphrasing. Embedding-based metrics capture meaning, but may smooth over small differences. To overcome the limitations of a single metric or method, we used multiple analytical tools and metrics in our study to obtain a more complete picture of how the agent-like interaction between two LLMs evolves. 

\section{Results and Discussion}
In this section, we report our findings. We performed both qualitative assessment of the output sentences for each round of the conversation. We also carried out extensive experiments to understand the convergence phenomenon quantitatively.

\subsection{Qualitative Analysis}
\begin{wrapfigure}{r}{0.5\textwidth}
    \centering
    \includegraphics[width=0.48\textwidth]{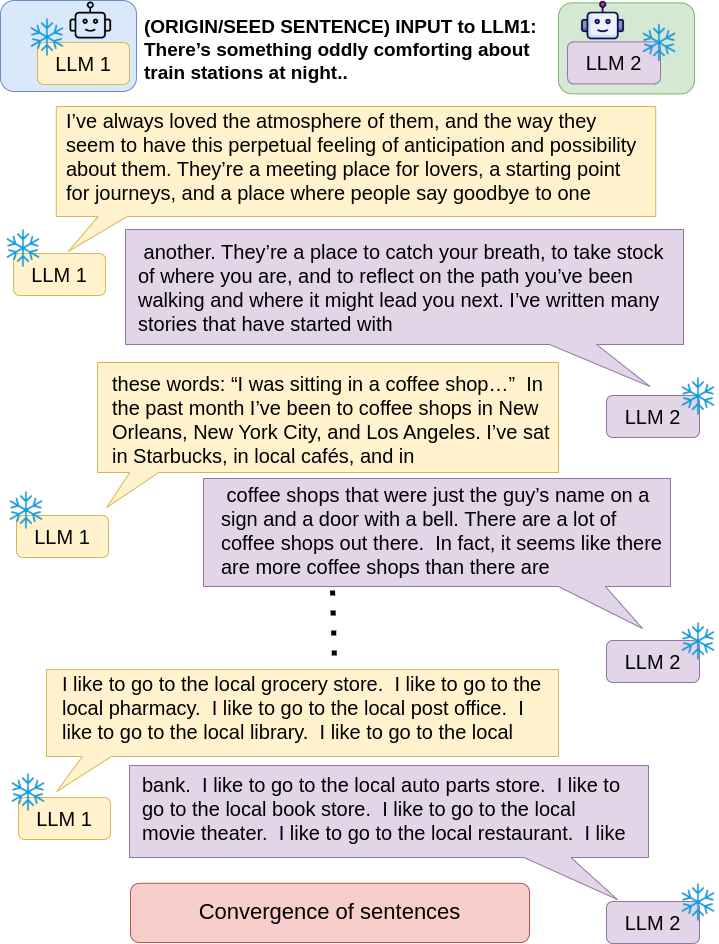}
    \caption{Example of convergence of the outputs of two models}
    \label{fig:convergence_example_figure}
\end{wrapfigure}

We found, irrespective of the source of the origin sentence (please refer to Table-\ref{tab:seed_sources}), the chosen model pairs maintain coherence for several steps before the first sign of convergence emerges. While there are some instances where the conversation continues and finishes $25$ turns (each turn has two steps corresponding to two model inferences in sequence) without convergence, the majority of the conversations converge and produce similar repetitive sentences after some steps. When such convergence (occurrence of repetition) occurs, both models reproduce each other's last output, producing loops that continues for many turns until the final turn (25th turn) occurs. This behavior suggests that once a phrase associated with low diversity is produced by one model, both models tend to reproduce the same output rather than diverge or provide a new direction in the conversation. As an example, we show in Fig. \ref{fig:convergence_example_figure} such a case where after just few turns, the models start producing similar sentences.

While the convergence phenomenon has been found for most of the cases, there are some cases, when the conversation continues without convergence. Our manual inspection found that out of 50 seed sentences, 35 led to convergence within 25 turns. The remaining 15 completed without entering a repetitive loop. The step at which convergence began varied widely, with some cases showing convergence early, and others continued until the final few turns. Topic drift was observed in certain non-converged runs. We found the conversation shifted significantly from the initial subject. Some shifts were found to be abrupt (such as a sudden appearance of code snippets, lists, or numerical data). 
\subsection{Quantitative Analysis}
To study the results quantitatively, we investigated how each generated sentence drifted from the previous sentence sequentially. We used four different metrics to quantify this drift. The quantification is done in different spaces such as lexical space, space of strings, embedding space etc. Four metrics have been used for this purpose as described in Section-\ref{sec:metrics}.

\textbf{Color Scheme Used to show Quantitative results: } In each of the following analysis and results, we used different colors to distinguish between the steps taken by the two different models. Steps taken by the same model are marked with the same color for a given round.

\subsubsection{Consistent convergence phenomenon using four different metrics: }
\begin{figure}[htbp]
\centering
\begin{minipage}[t]{0.48\textwidth}
    \centering
    \includegraphics[width=\linewidth]{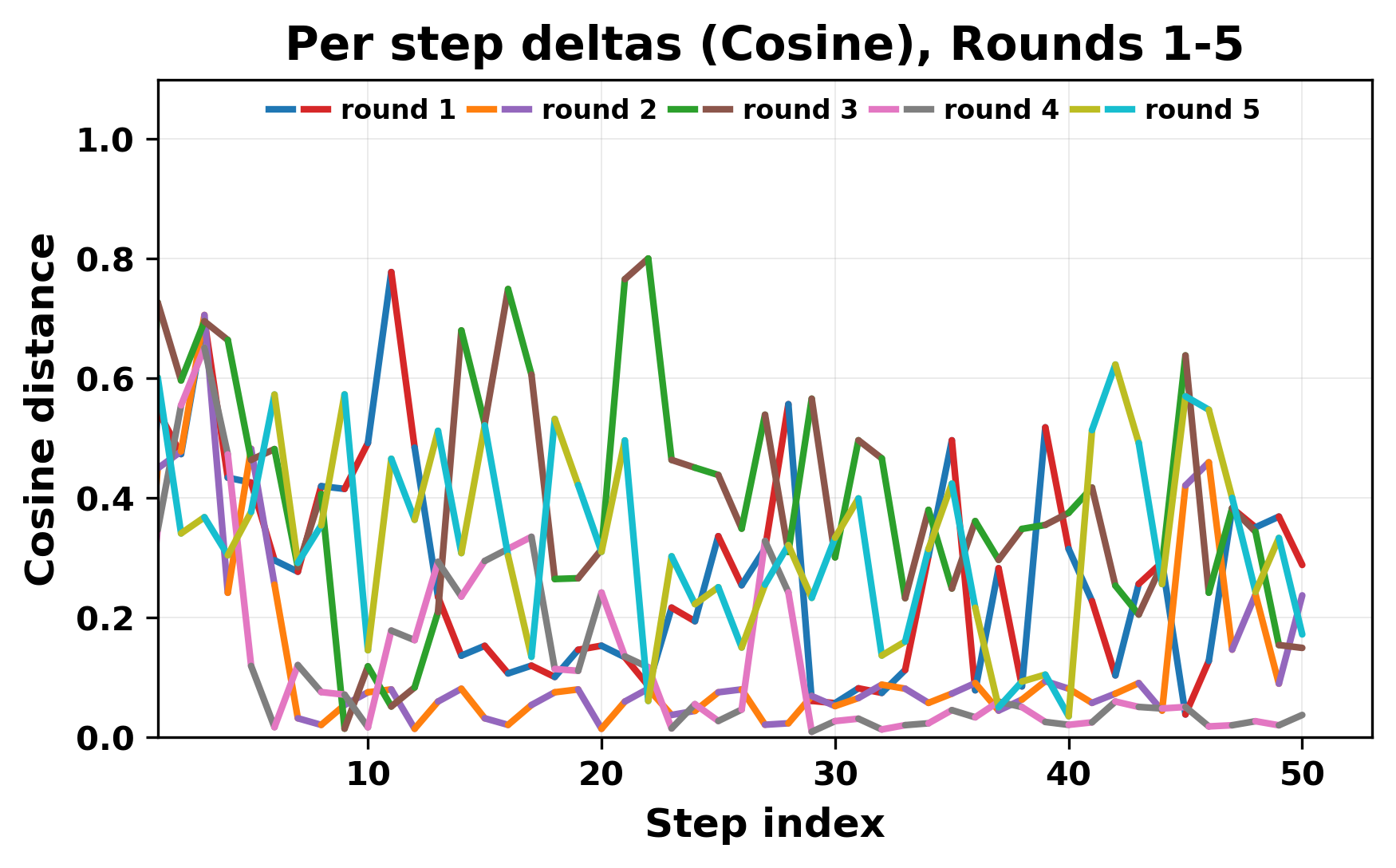}
    \includegraphics[width=\linewidth]{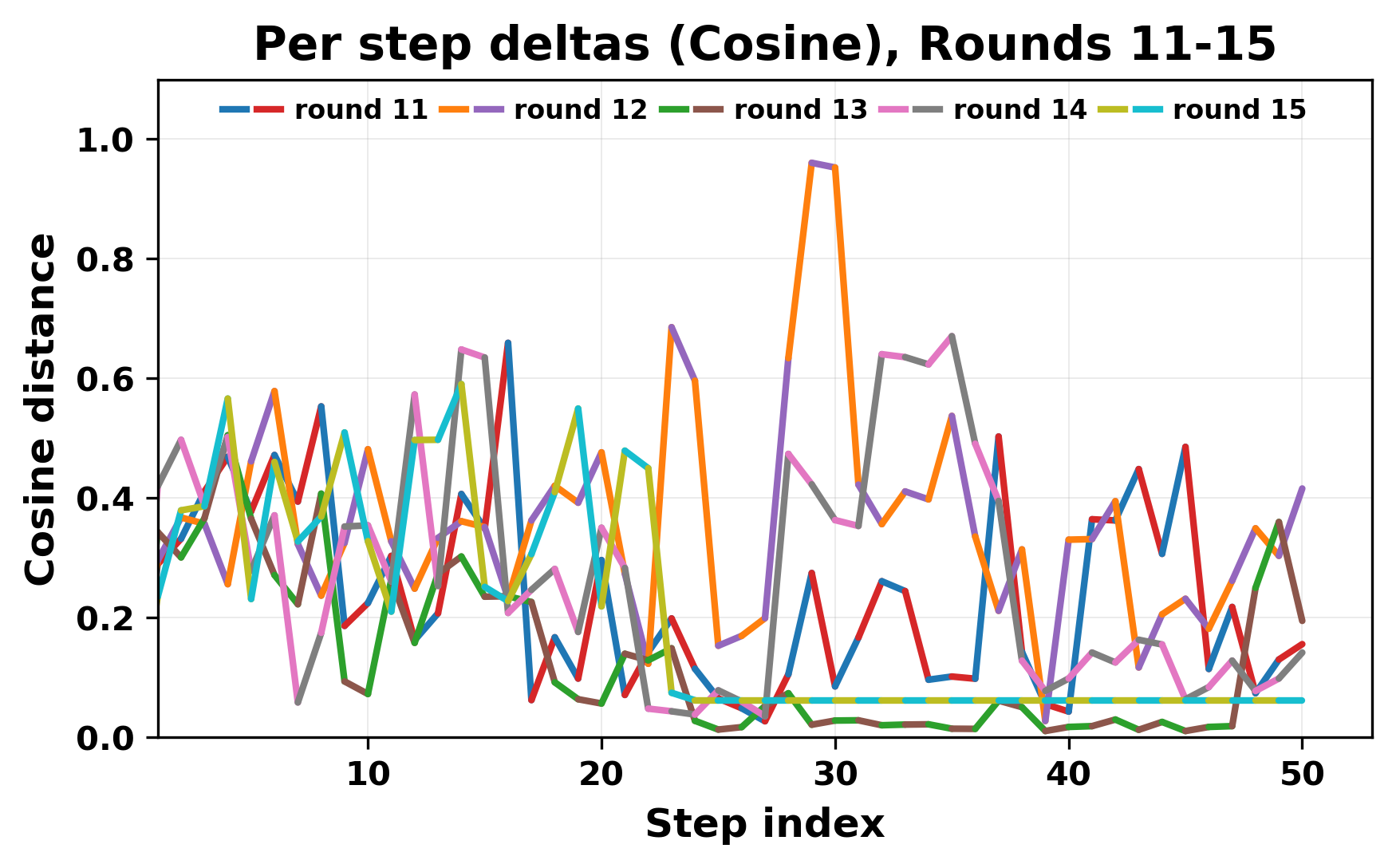}
    \includegraphics[width=\linewidth]{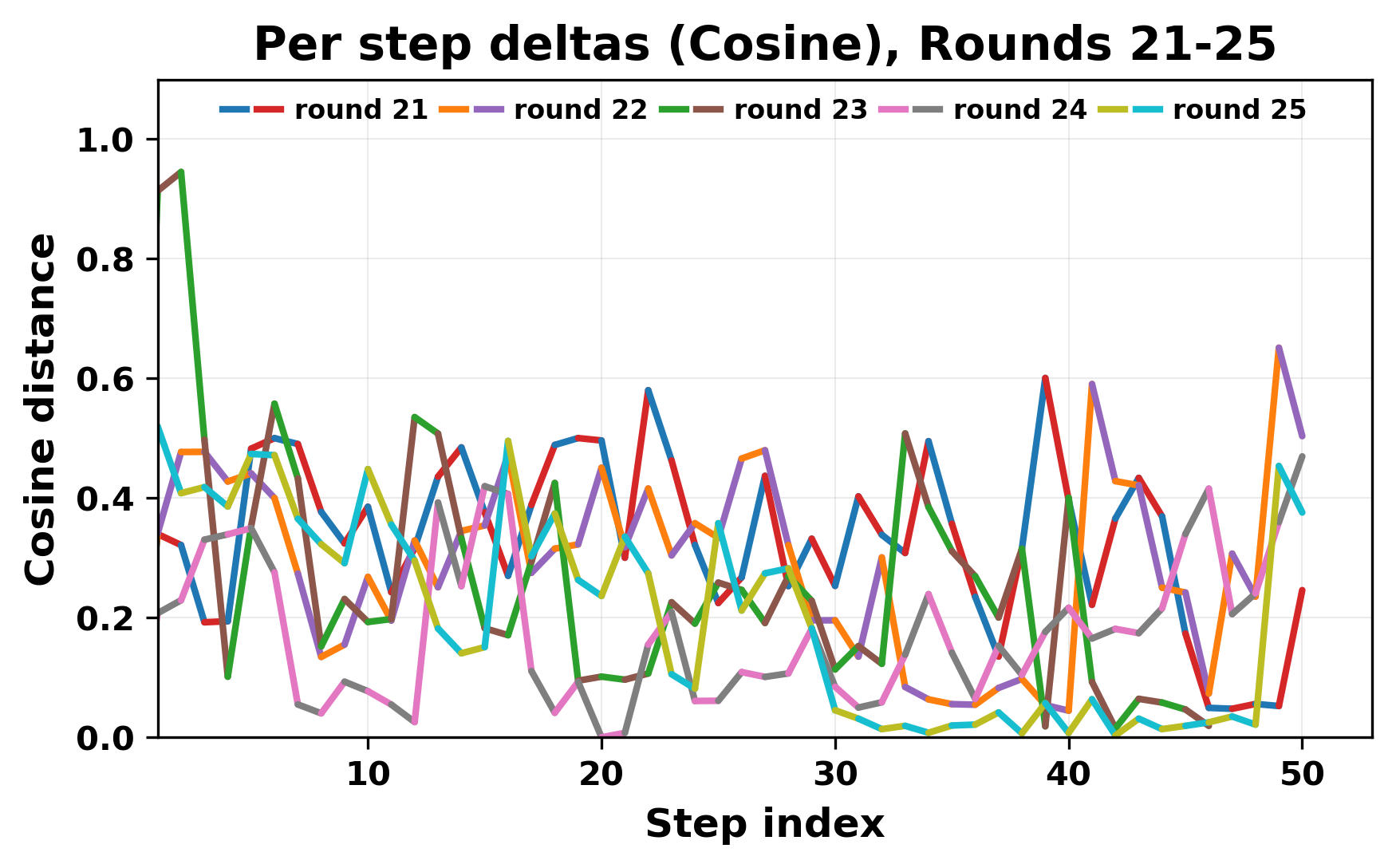}
    \includegraphics[width=\linewidth]{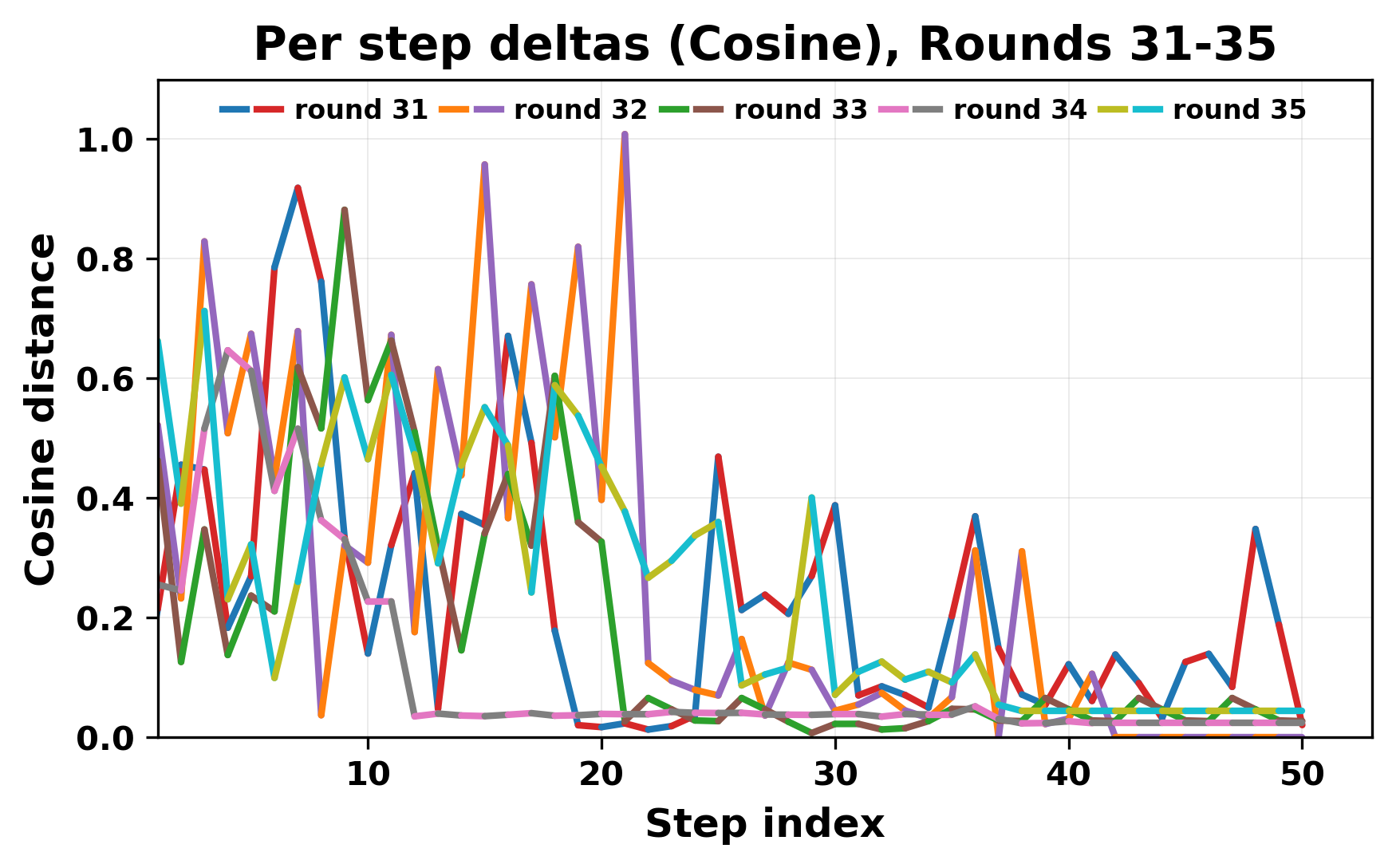}
    \includegraphics[width=\linewidth]{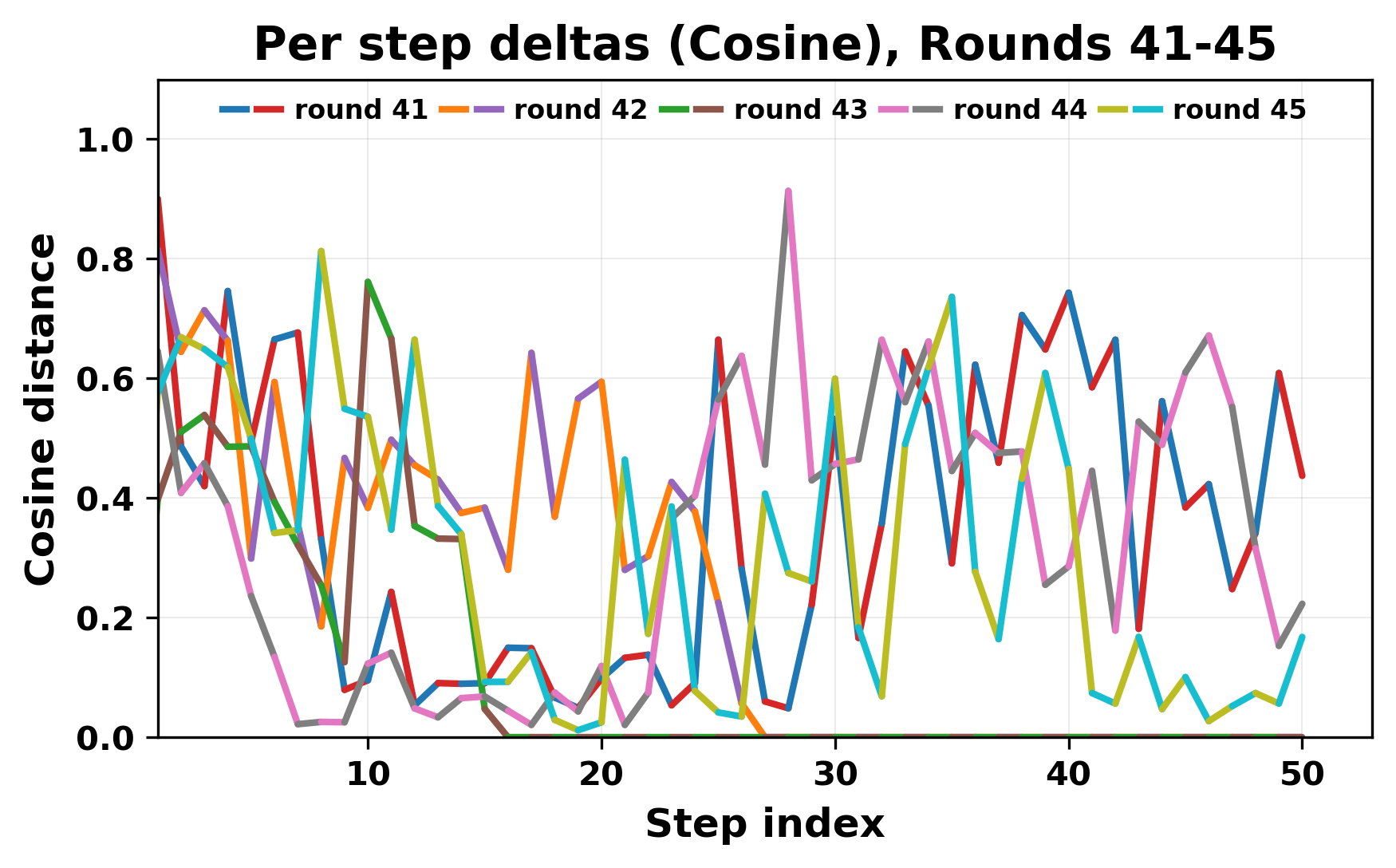}
\end{minipage}%
\hfill
\begin{minipage}[t]{0.46\textwidth}
    \centering
    \includegraphics[width=\linewidth]{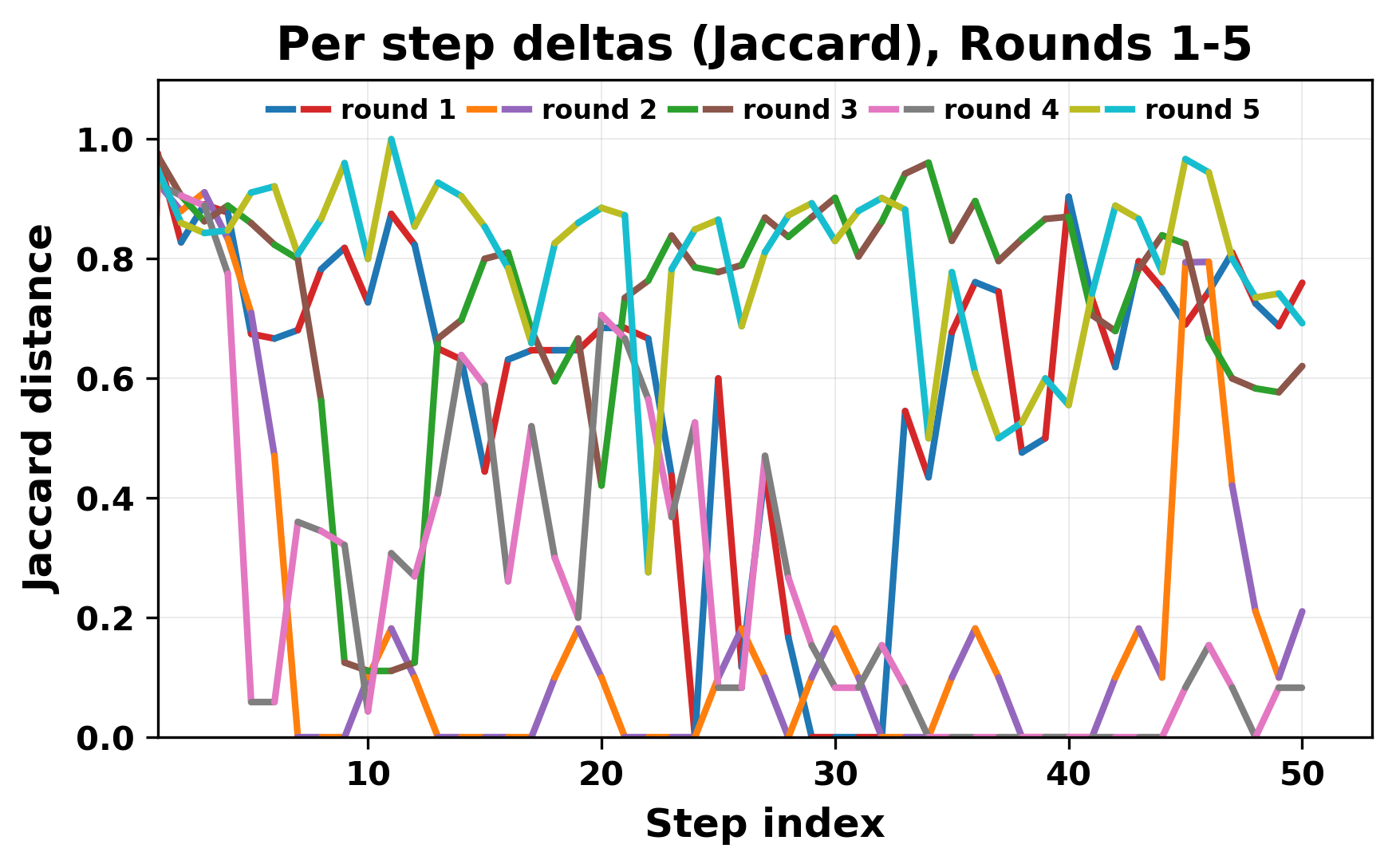}
    \includegraphics[width=\linewidth]{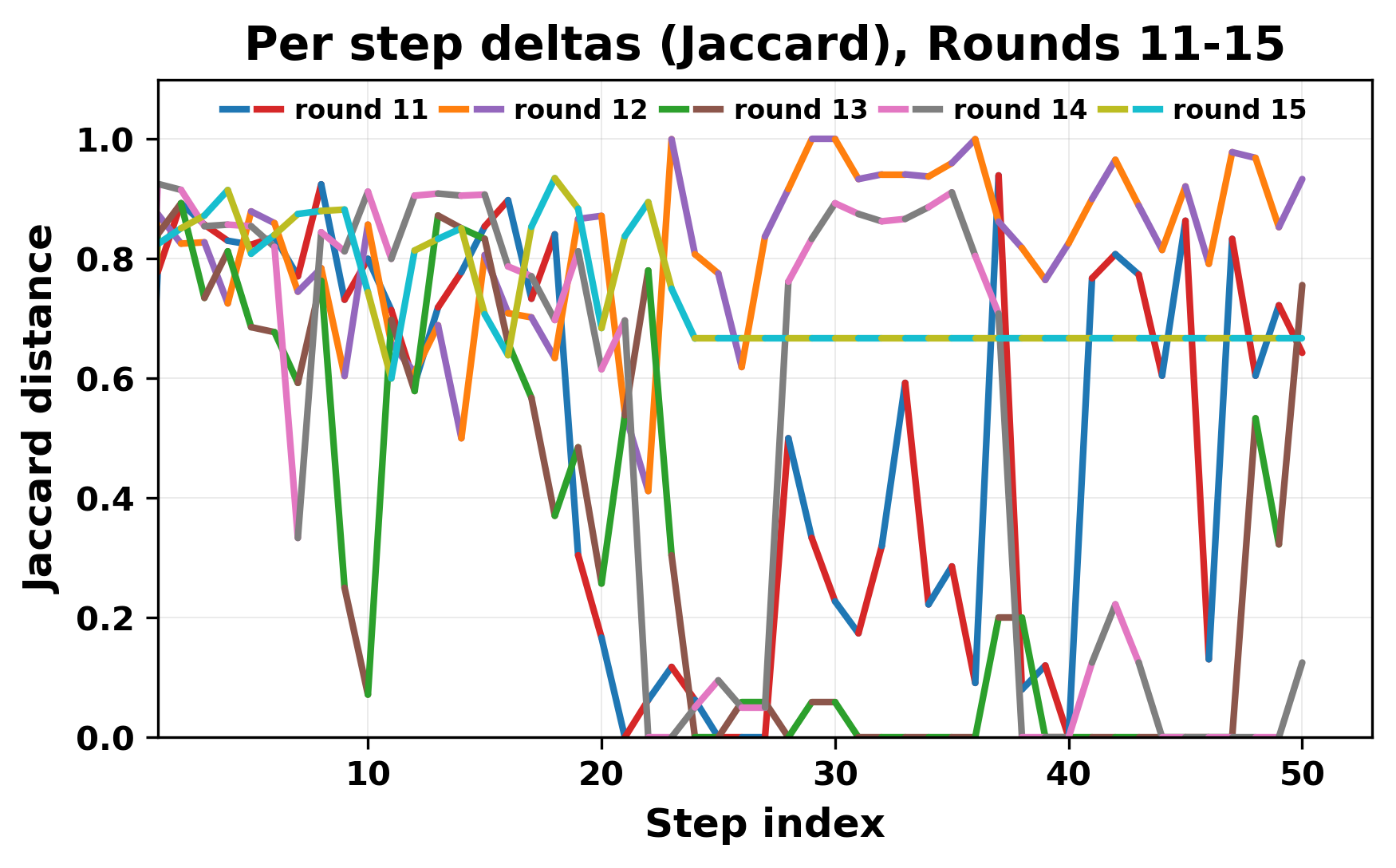}
    \includegraphics[width=\linewidth]{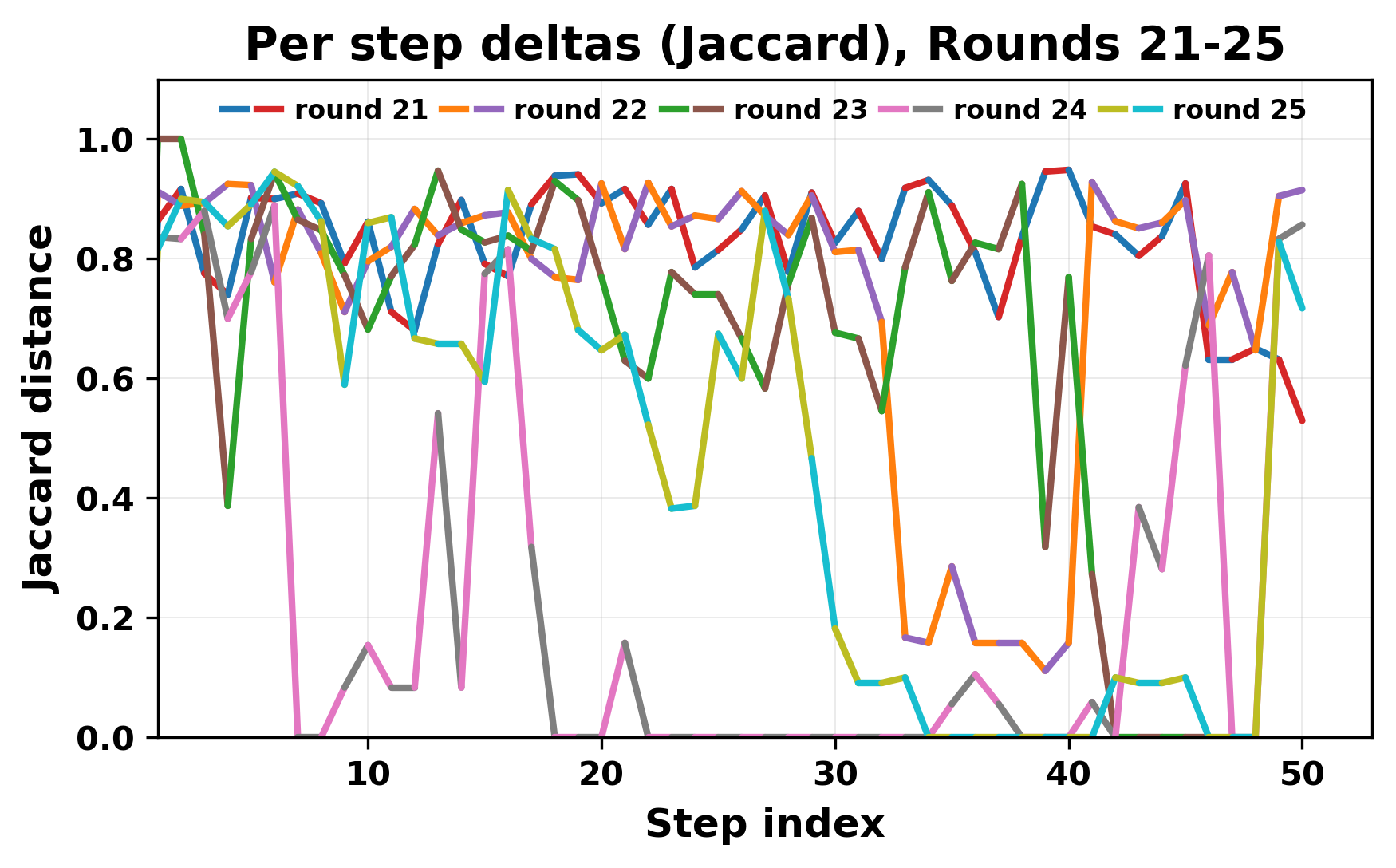}
    \includegraphics[width=\linewidth]{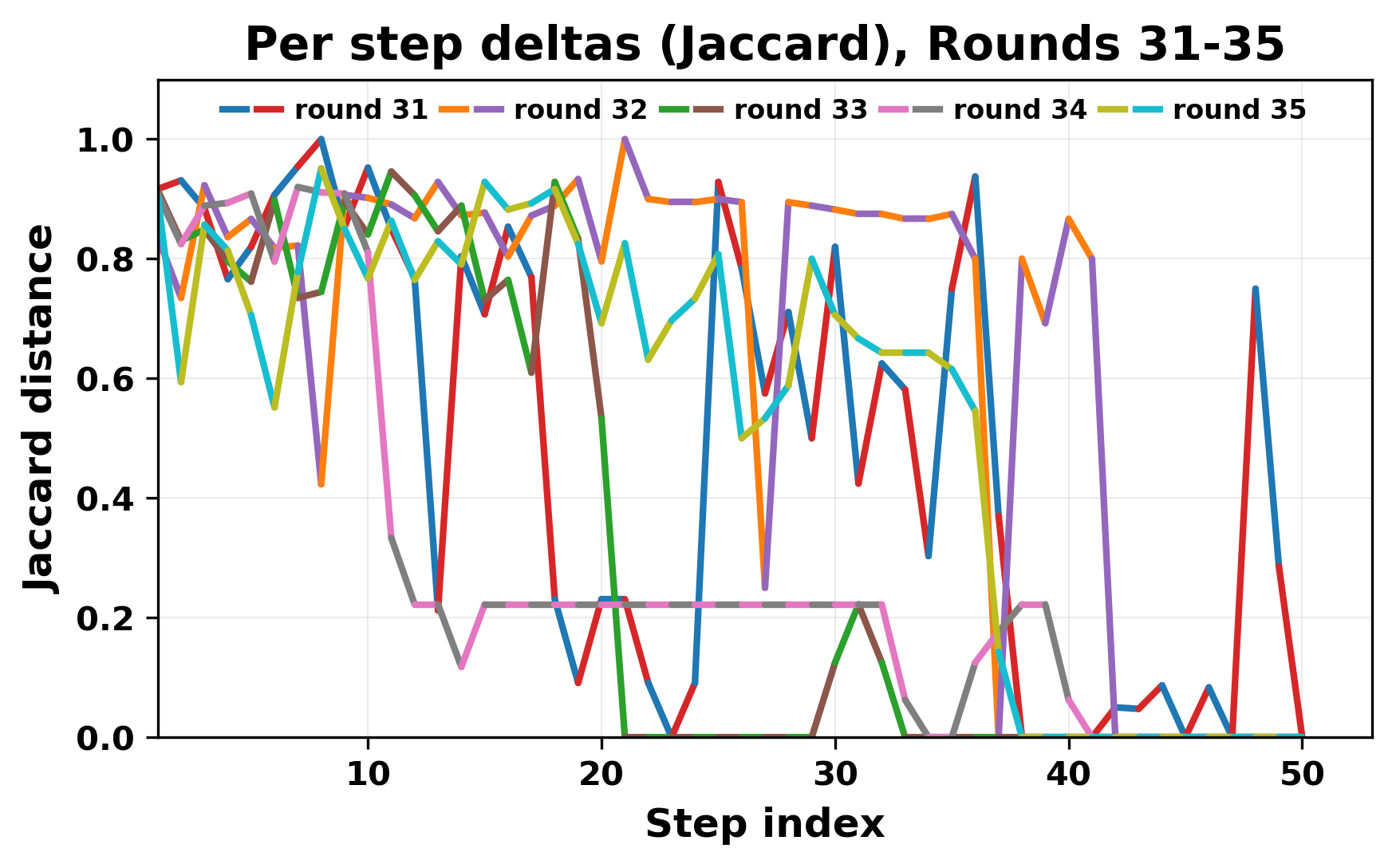}
    \includegraphics[width=\linewidth]{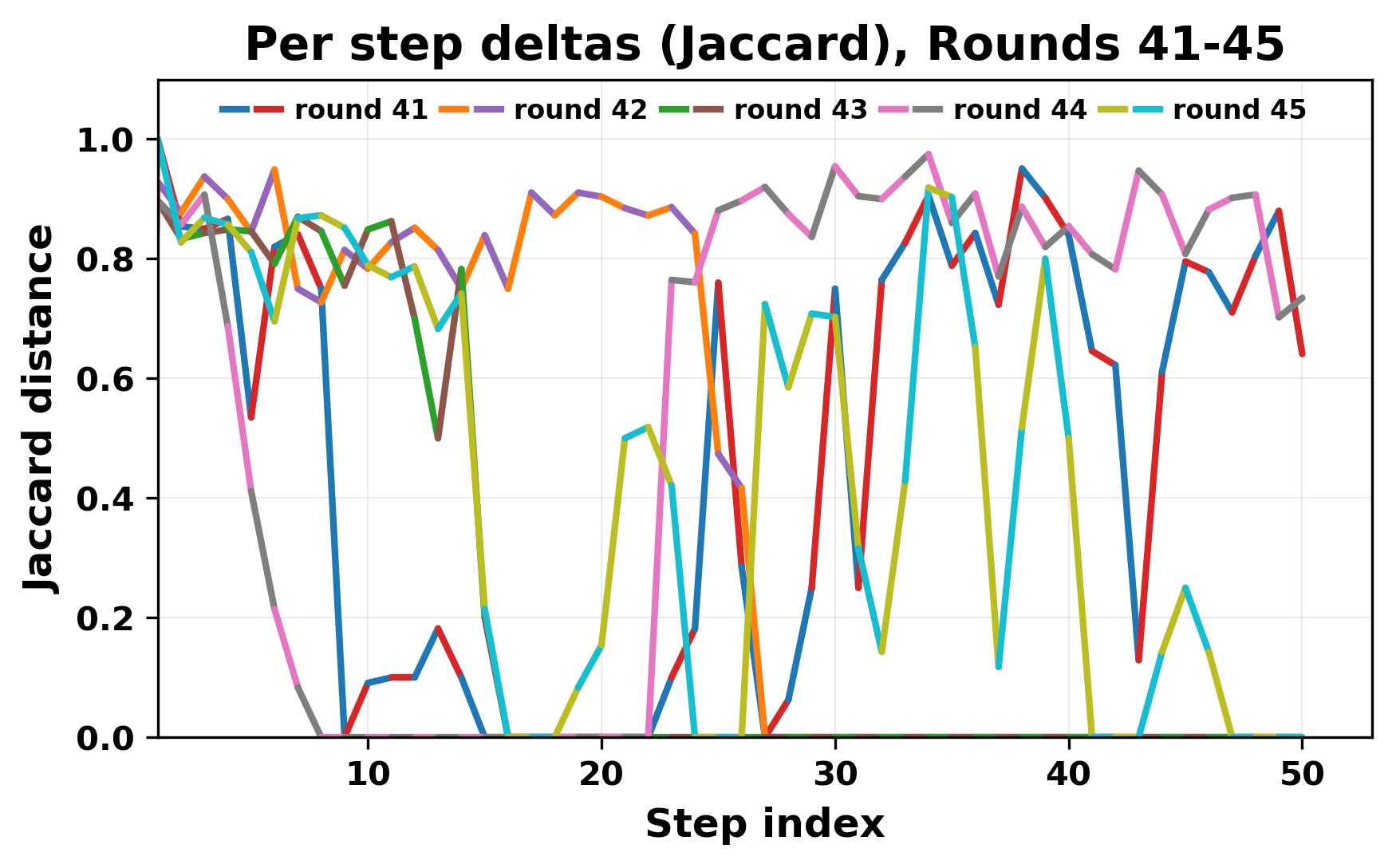}
\end{minipage}
\caption{Left: Cosine Distance; Right: Jaccard Distance}
\label{fig:cosine_jaccard_all}
\end{figure}

In Fig.\ref{fig:cosine_jaccard_all}, we showed  how the distance between two consecutive steps varies from turn 1 to turn 25 using Cosine Distance in the Embedding Space (Left) and Jaccard Distance (Right). We see example of convergence in each of these figures. Convergence happens when the distance between consecutive turns drops significantly. In later section, we developed an automated cut-off based approach to detect these convergences automatically.

\begin{figure}[hbtp]
\centering
\begin{minipage}[t]{0.48\textwidth}
    \centering
    \includegraphics[width=\linewidth]{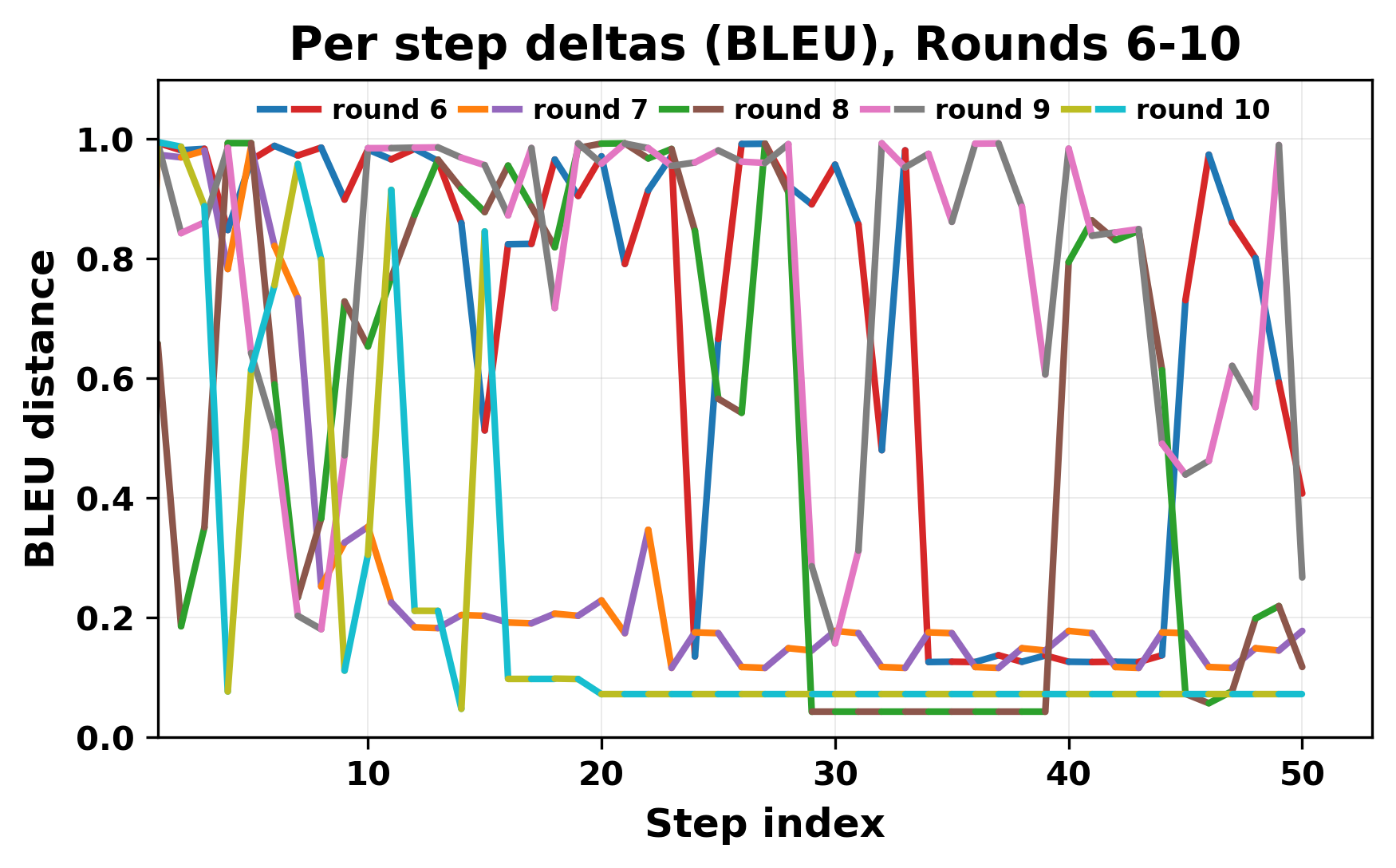}
    \includegraphics[width=\linewidth]{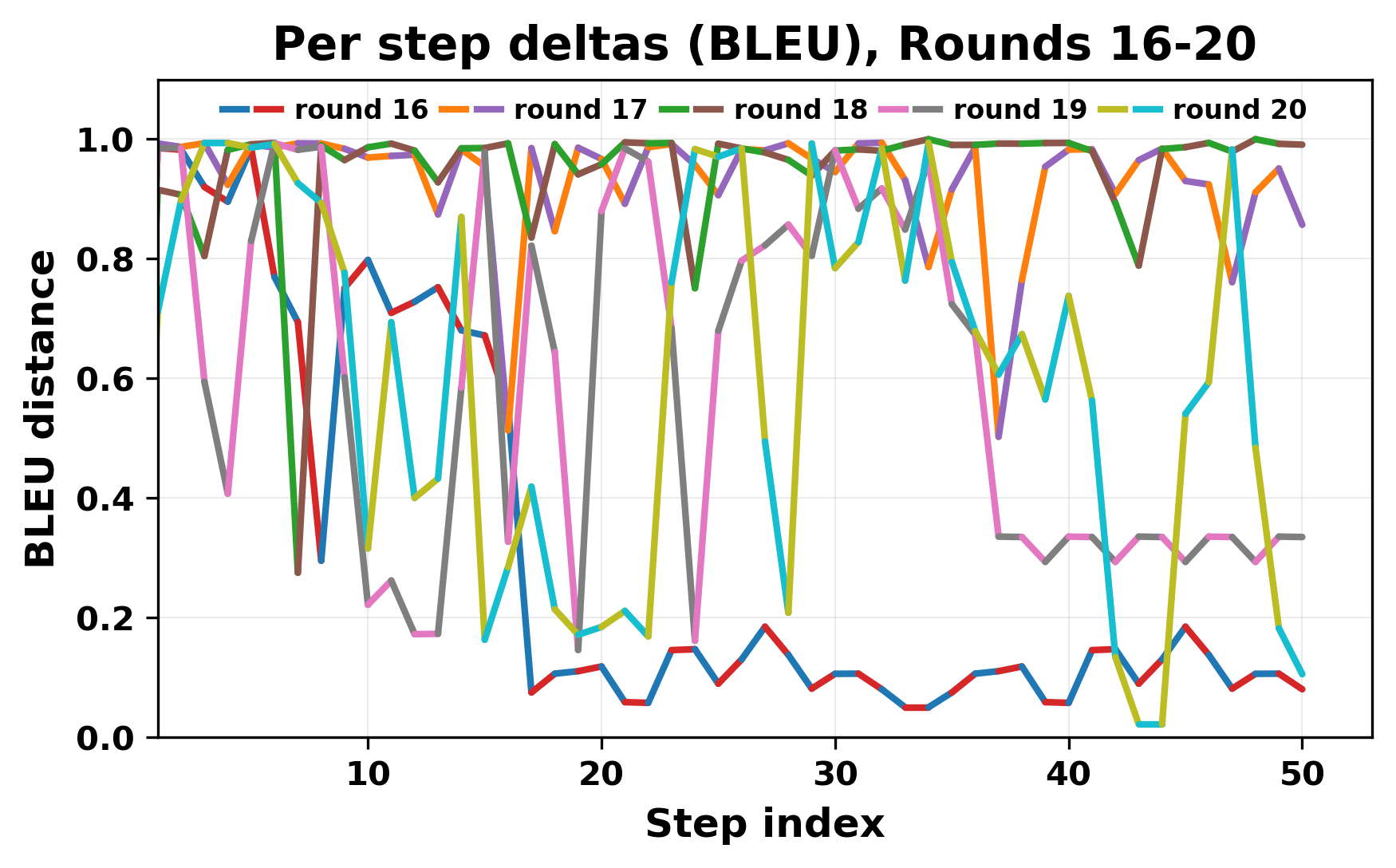}
    \includegraphics[width=\linewidth]{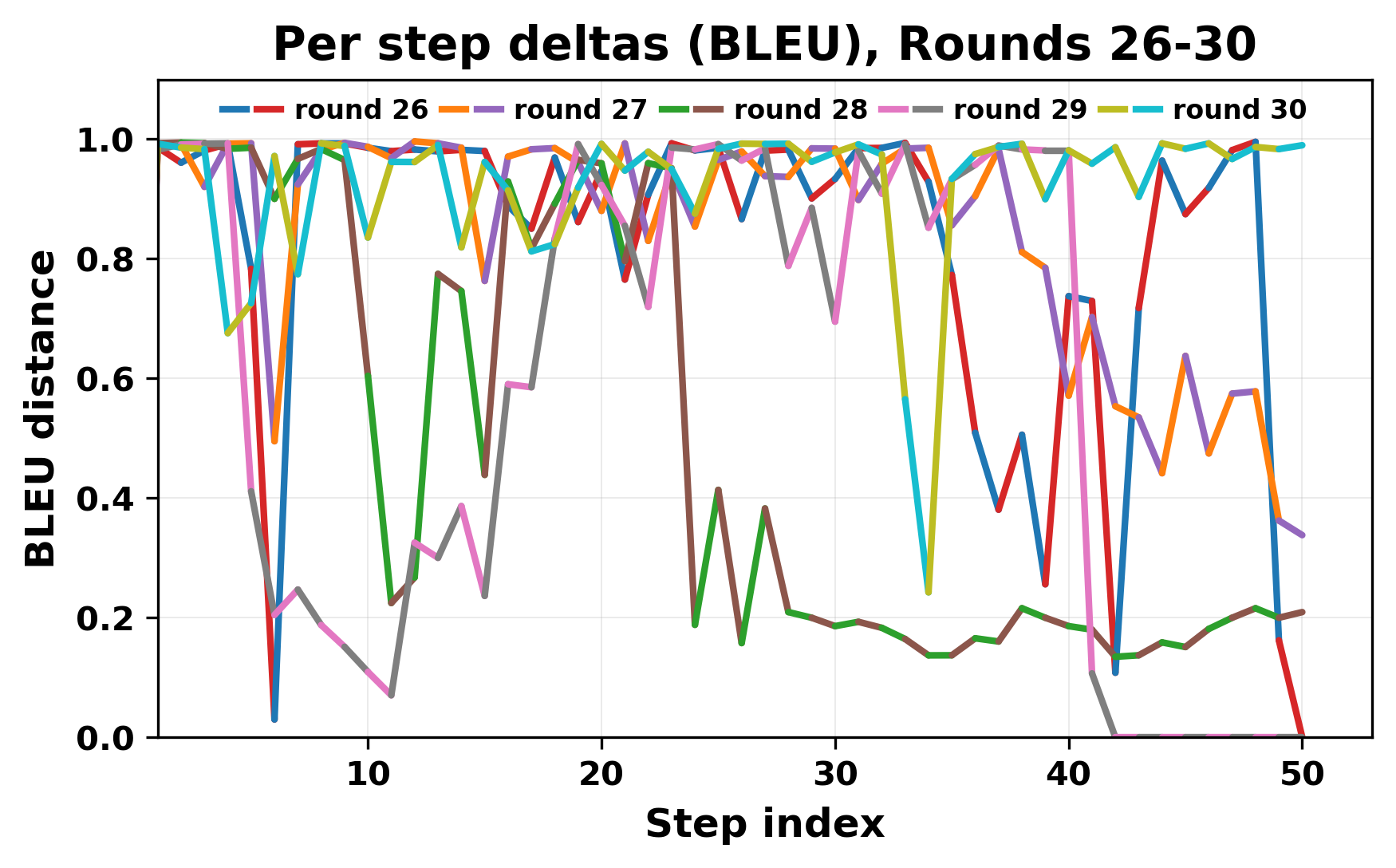}
    \includegraphics[width=\linewidth]{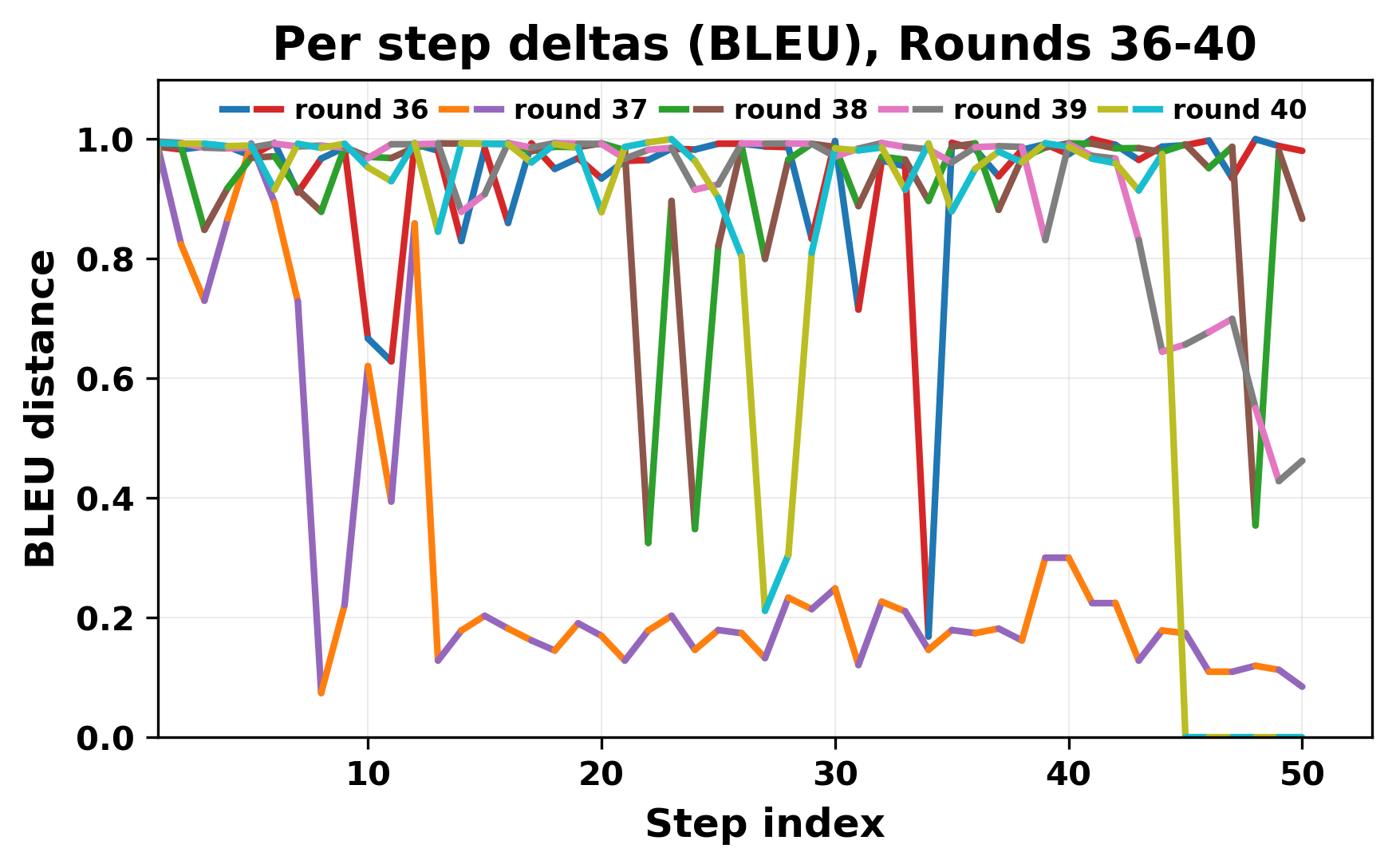}
    \includegraphics[width=\linewidth]{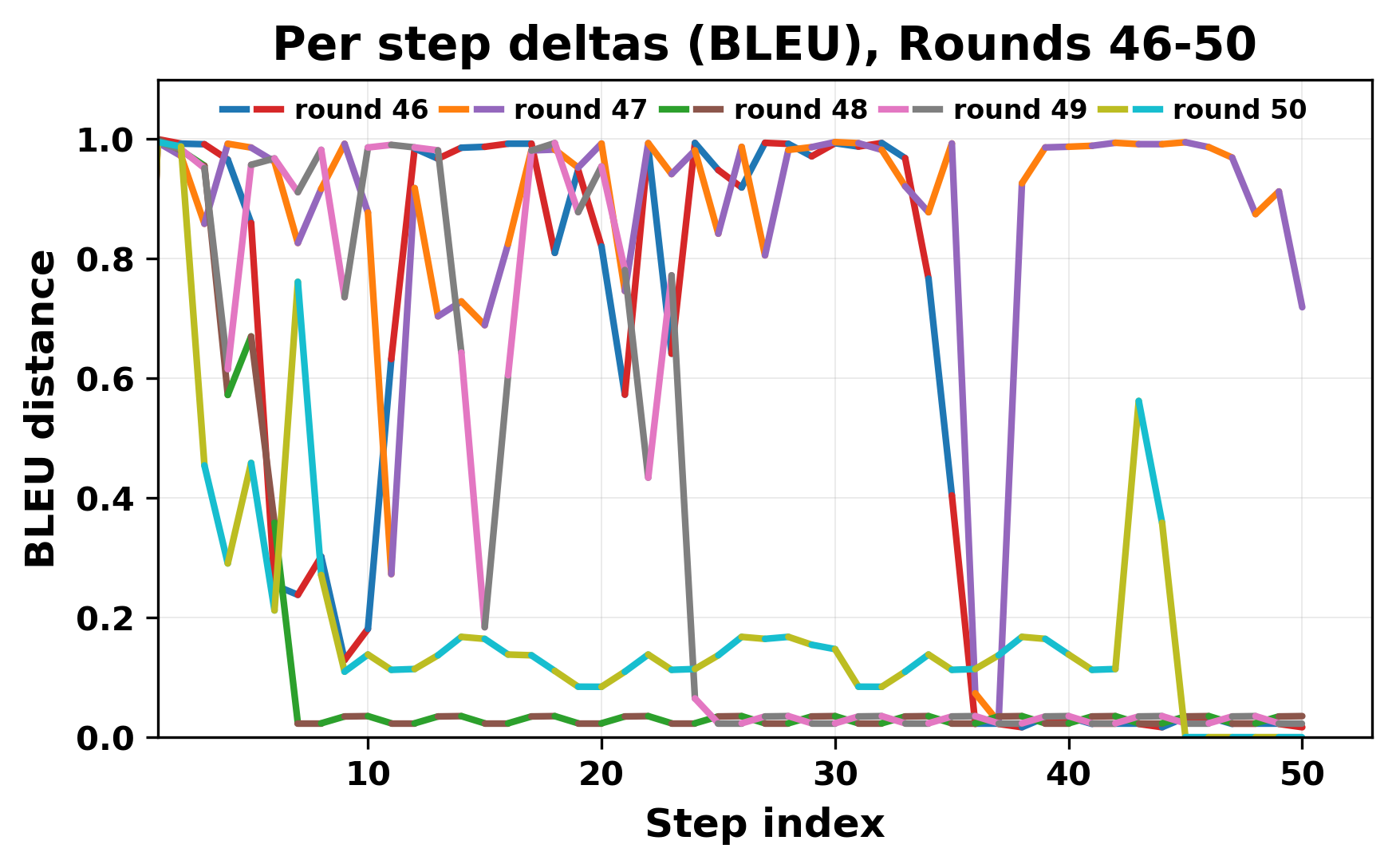}
\end{minipage}
\hfill
\begin{minipage}[t]{0.46\textwidth}
    \centering
    \includegraphics[width=\linewidth]{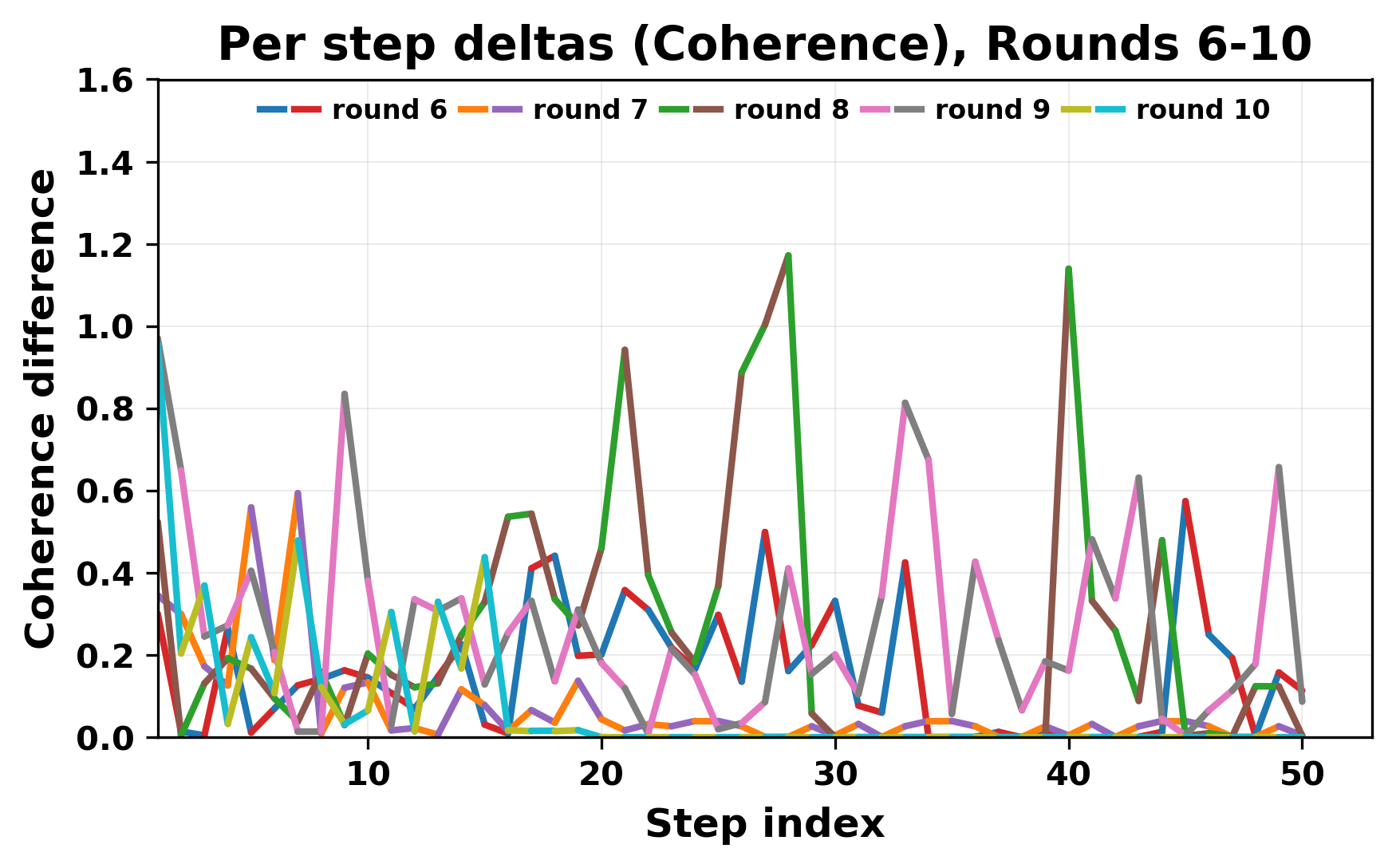}
    \includegraphics[width=\linewidth]{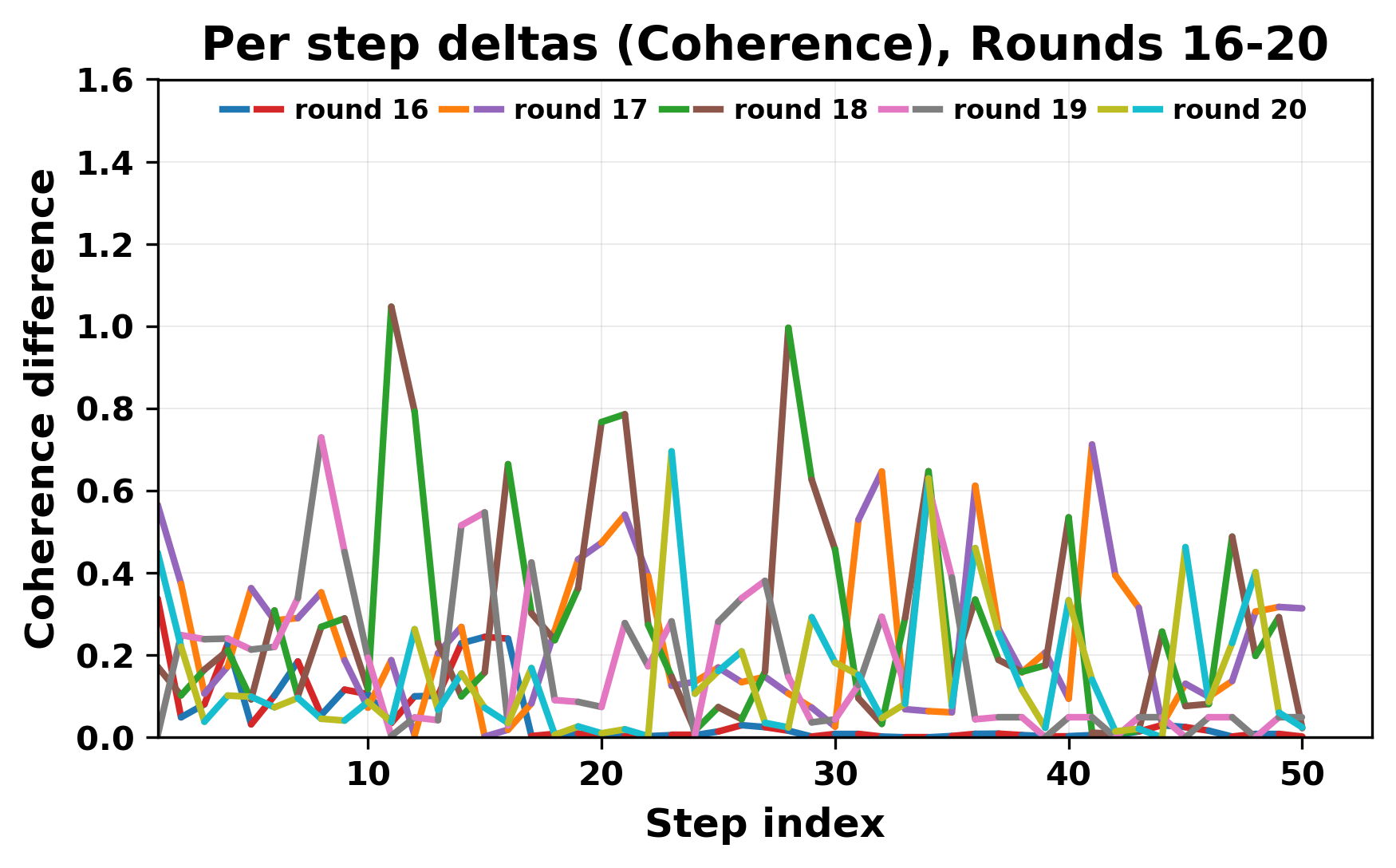}
    \includegraphics[width=\linewidth]{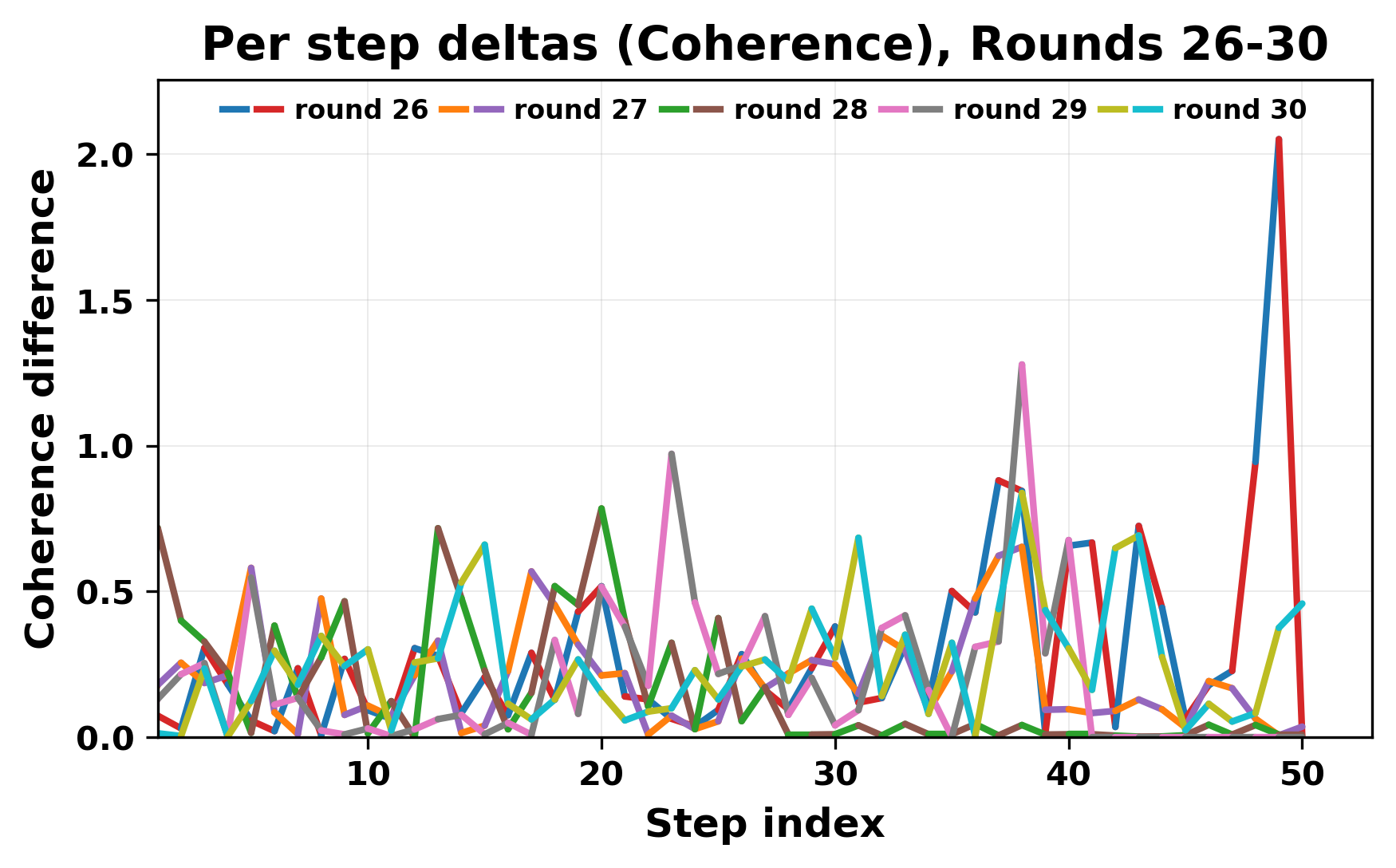}
    \includegraphics[width=\linewidth]{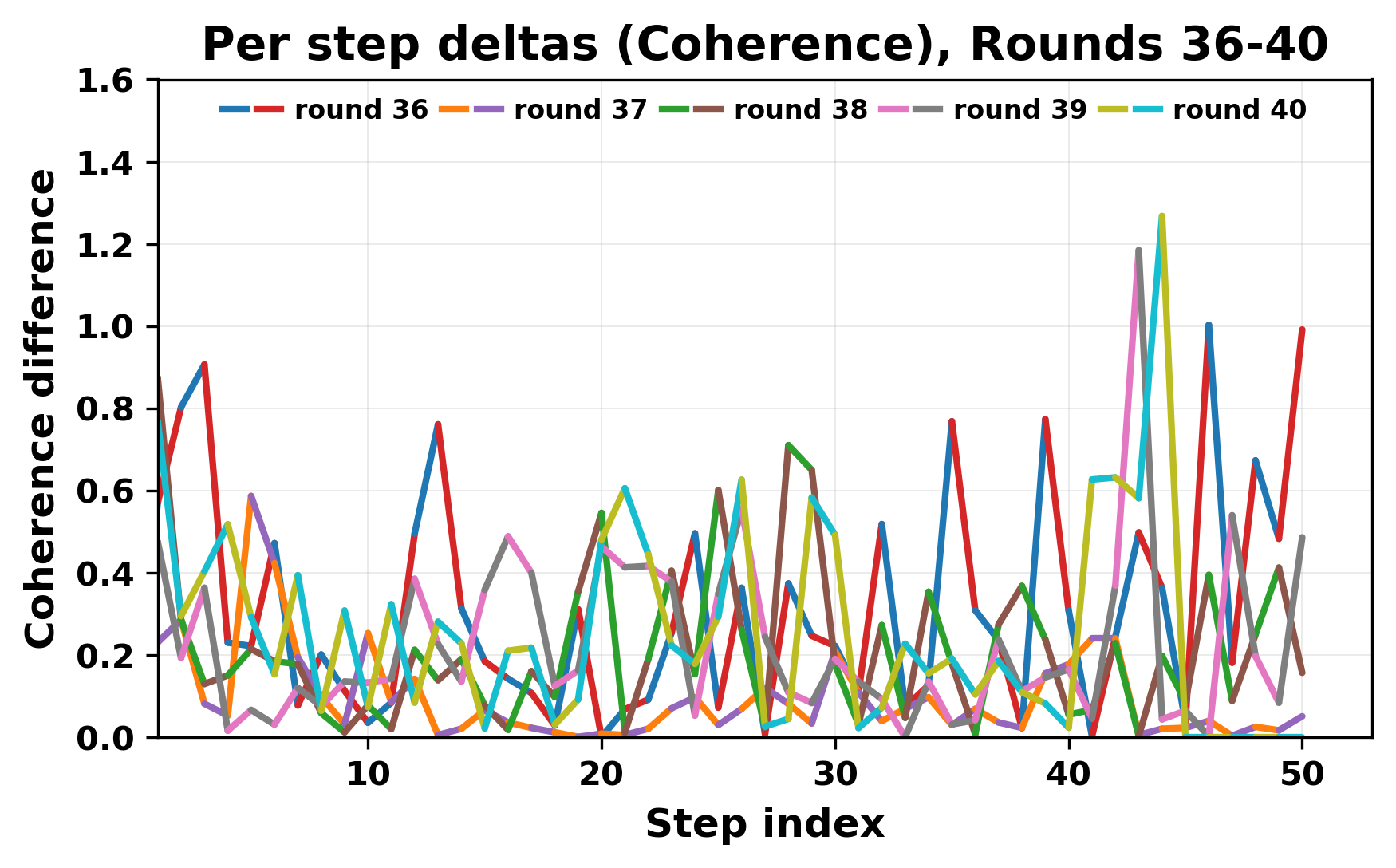}
    \includegraphics[width=\linewidth]{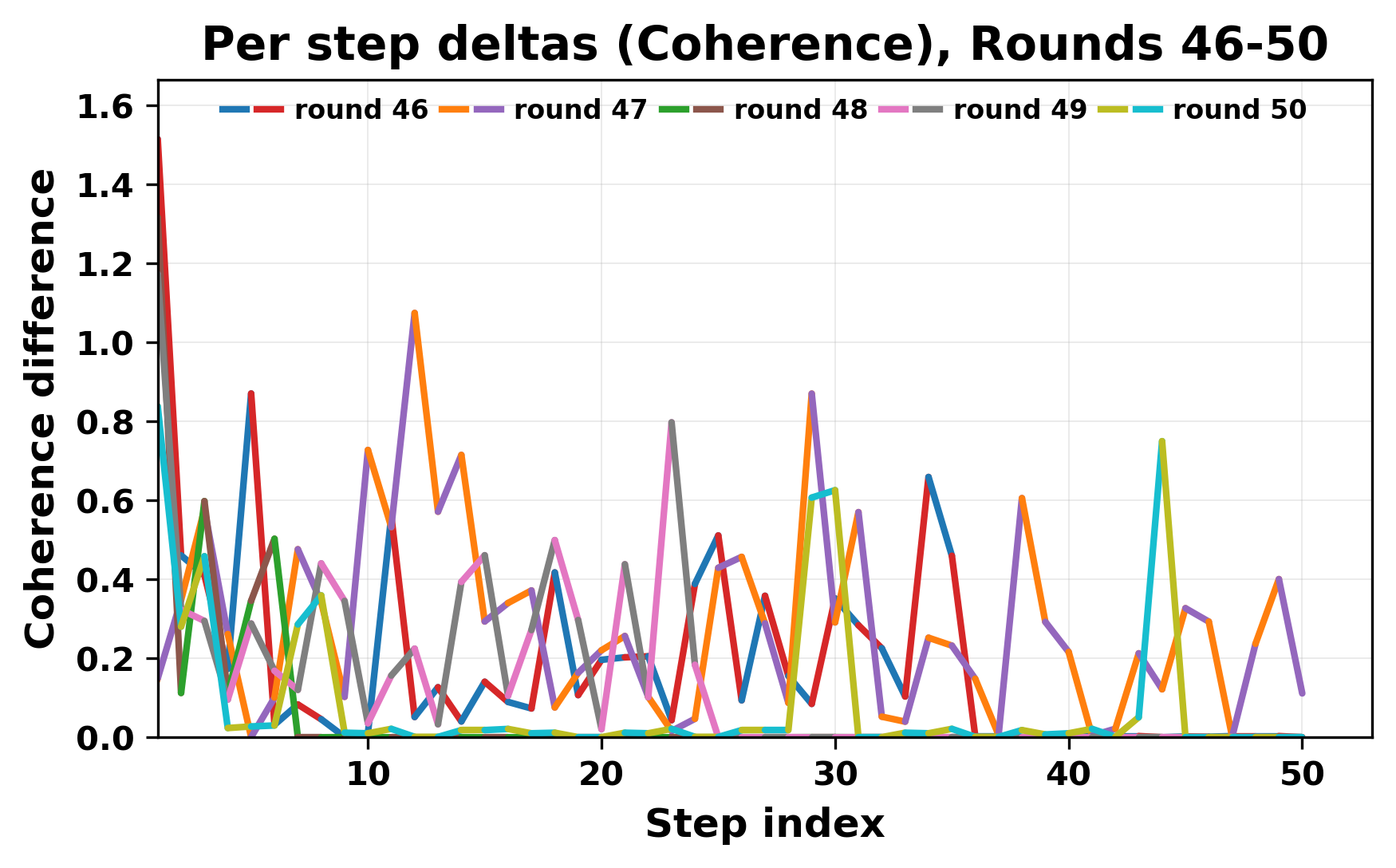}
\end{minipage}
\caption{Left: BLEU score-based distance; Right: Coherence}
\label{fig:bleu_coherence_all}
\end{figure}

Similarly, in Fig.\ref{fig:bleu_coherence_all}, we showed BLEU score based distance between two consecutive steps from turn 1 to turn 25 (Left) and the same using Coherence (Right). 

In each of these rounds, we had different seed sentences selected from 5 different sources (refer to Table-\ref{tab:seed_sources}). Using any of the selected distance metric, there are clear signs of convergence in many conversations irrespective of the source of the seed sentence.

\subsubsection{Embedding Projection and t-SNE Visualization: }

To visualize how model outputs evolve over time in embedding space, we projected the embeddings of each output onto two dimensions using t-distributed stochastic neighbor embedding (t-SNE). 

\begin{figure}[htbp]
\centering
\begin{minipage}[t]{0.50\textwidth}
    \centering
    \includegraphics[width=\linewidth]{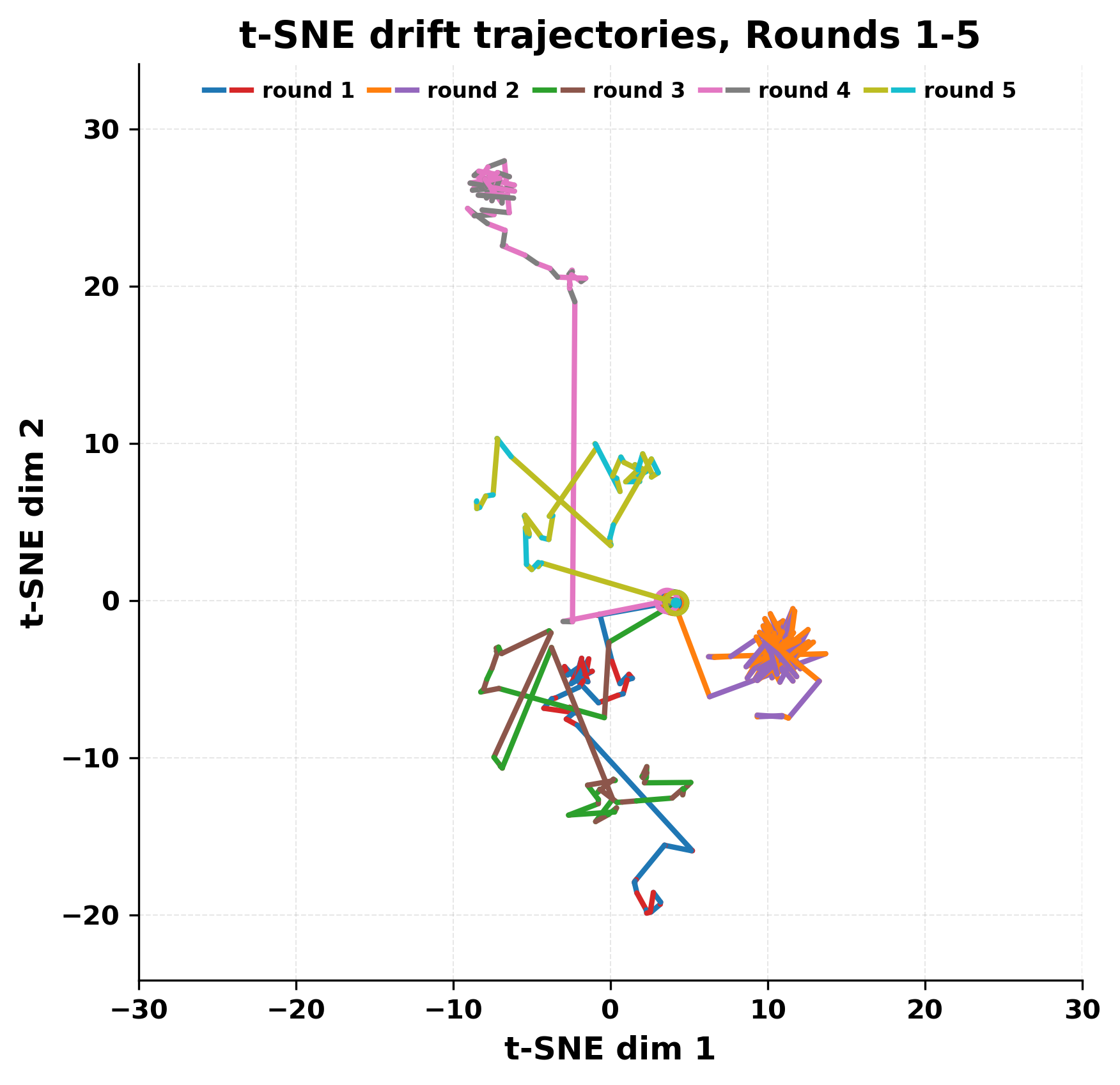}
    
    \includegraphics[width=\linewidth]{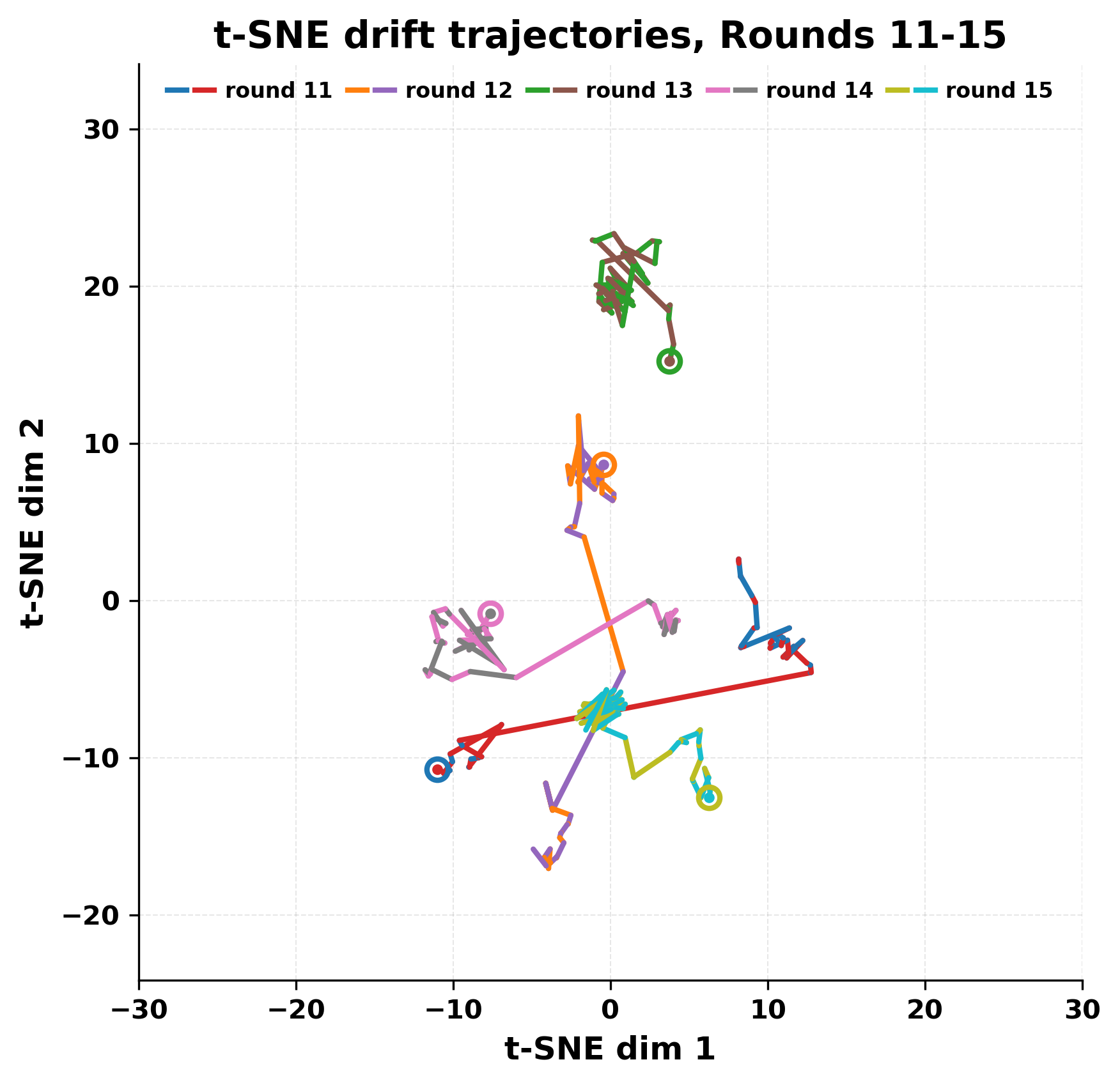}
    
    \includegraphics[width=\linewidth]{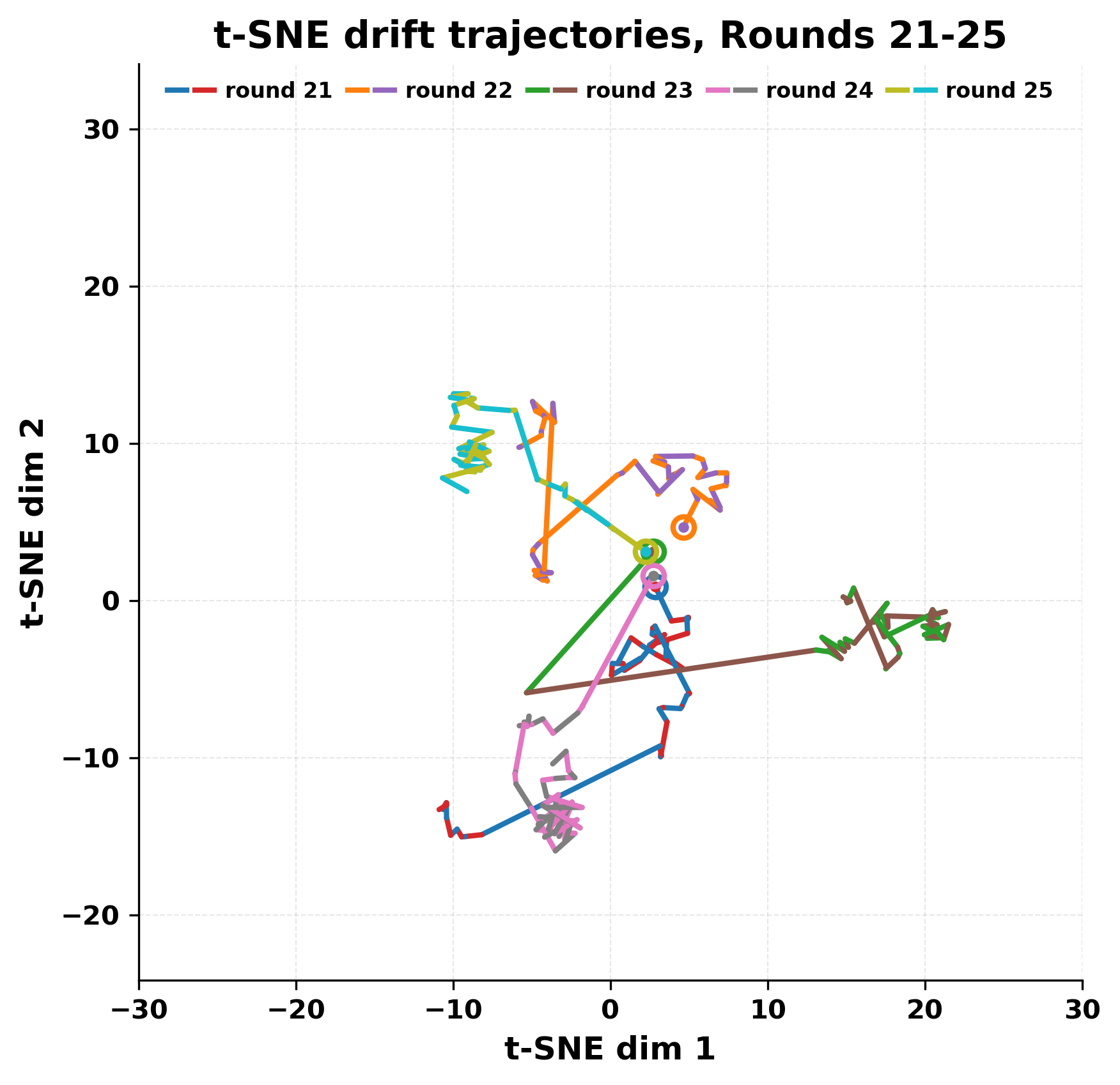}
\end{minipage}%
\hfill
\begin{minipage}[t]{0.50\textwidth}
    \centering
    \includegraphics[width=\linewidth]{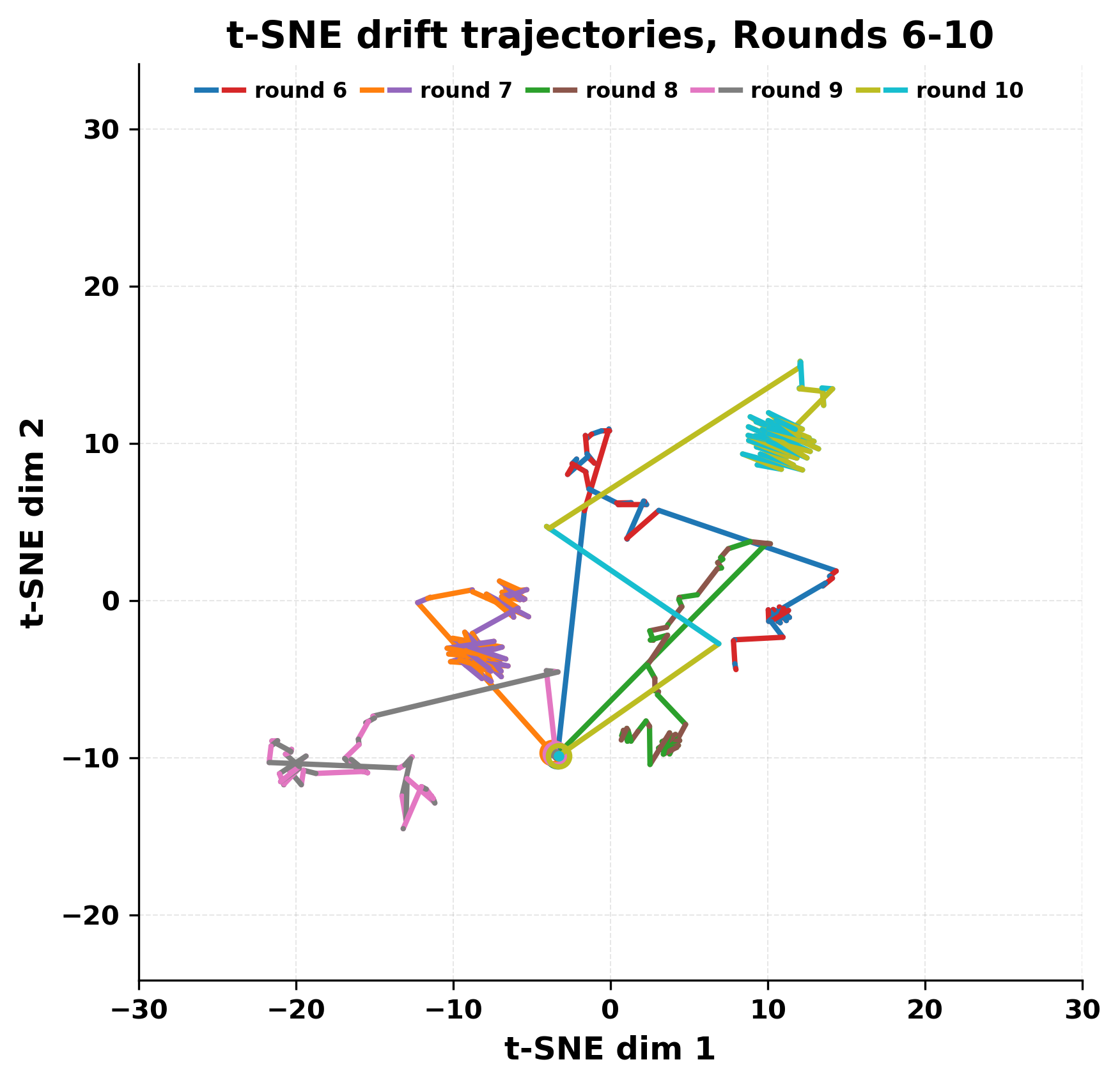}
    
    \includegraphics[width=\linewidth]{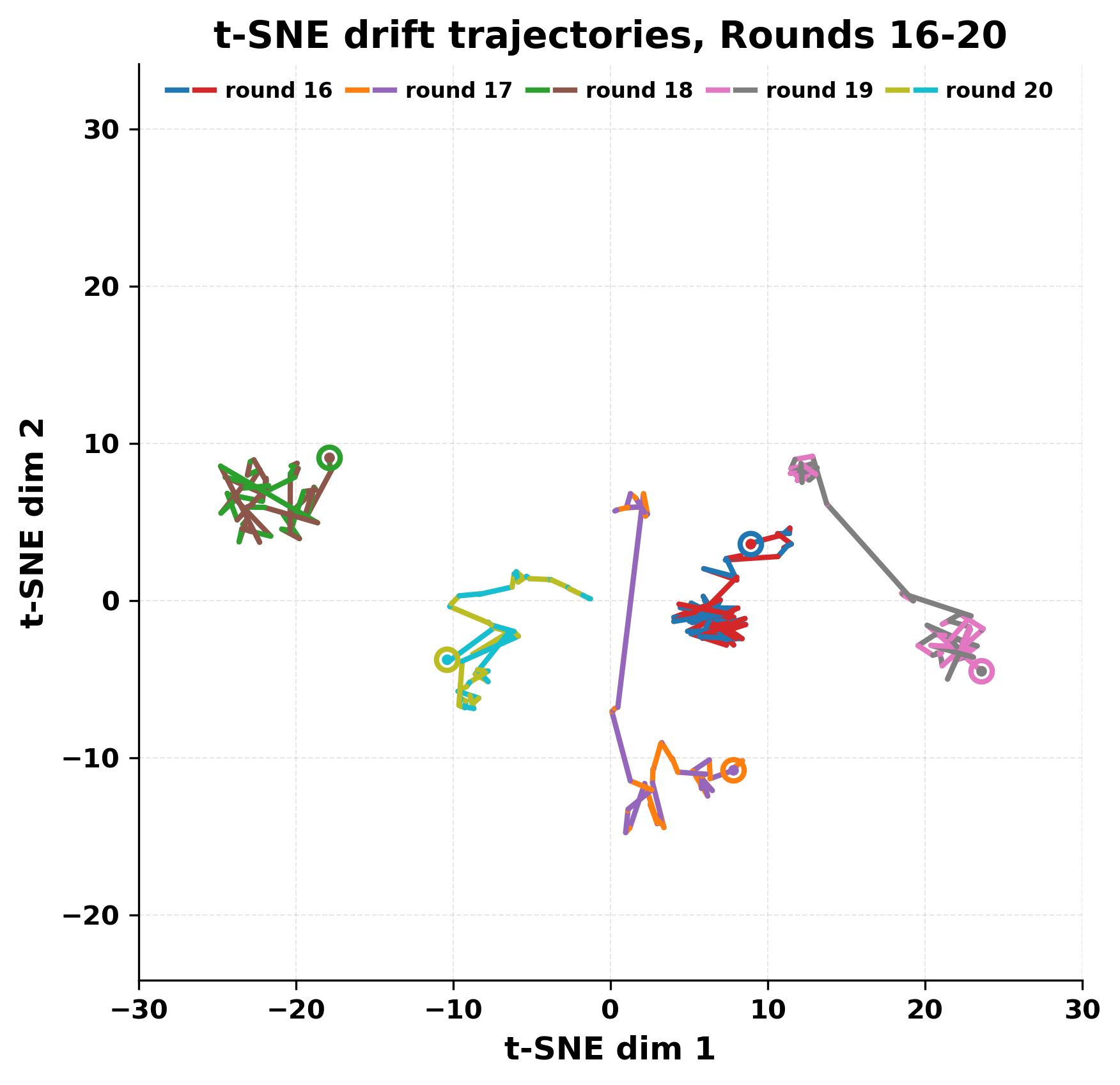}
    
    \includegraphics[width=\linewidth]{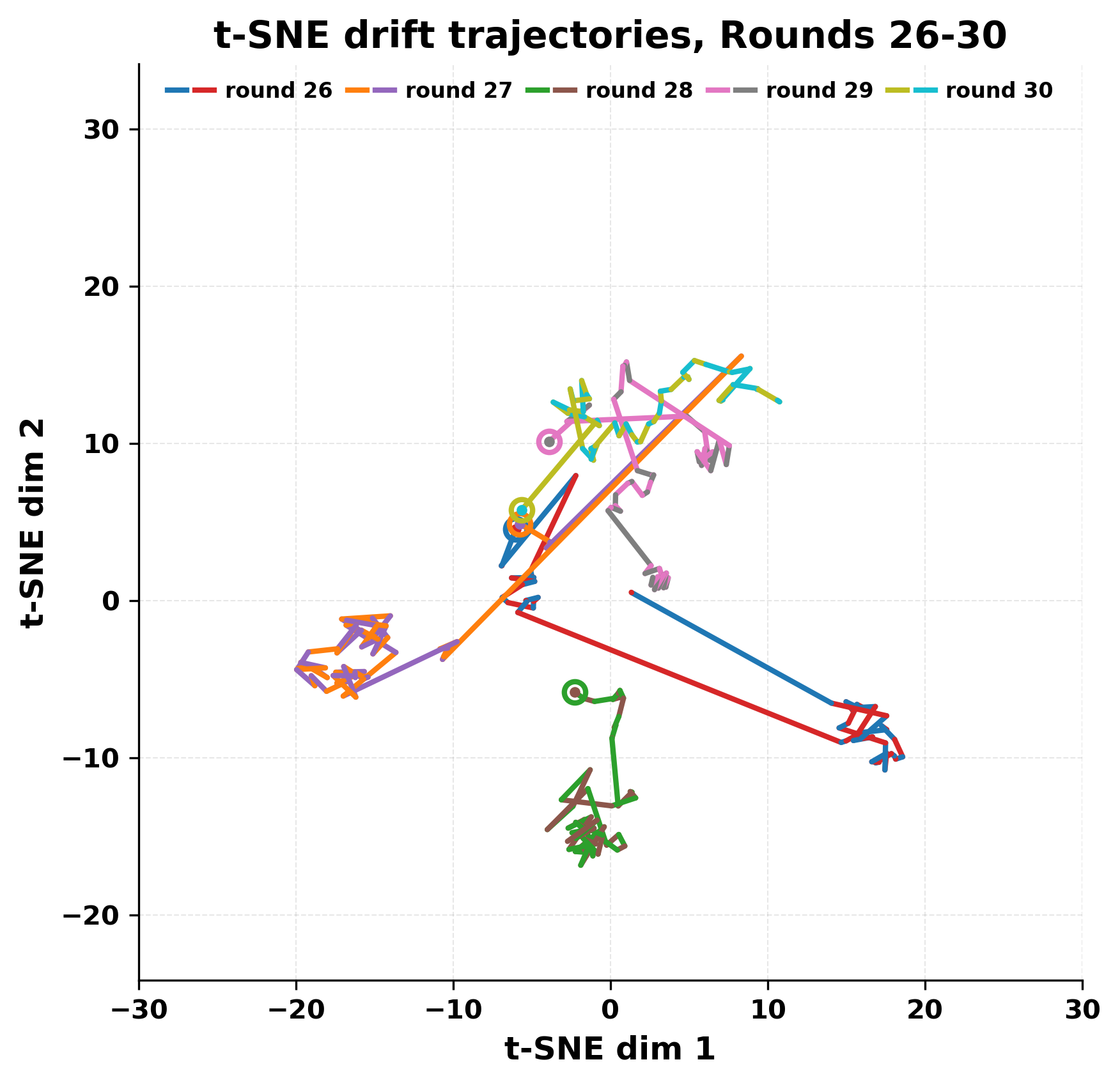}
\end{minipage}
\caption{t-SNE projection of sentence embeddings for Rounds 1–30, grouped in intervals of 5. Each point is a model output.}
\label{fig:tsne1}
\end{figure}

\begin{figure}[htbp]
\centering

\begin{minipage}[t]{0.50\textwidth}
    \centering
    \includegraphics[width=\linewidth]{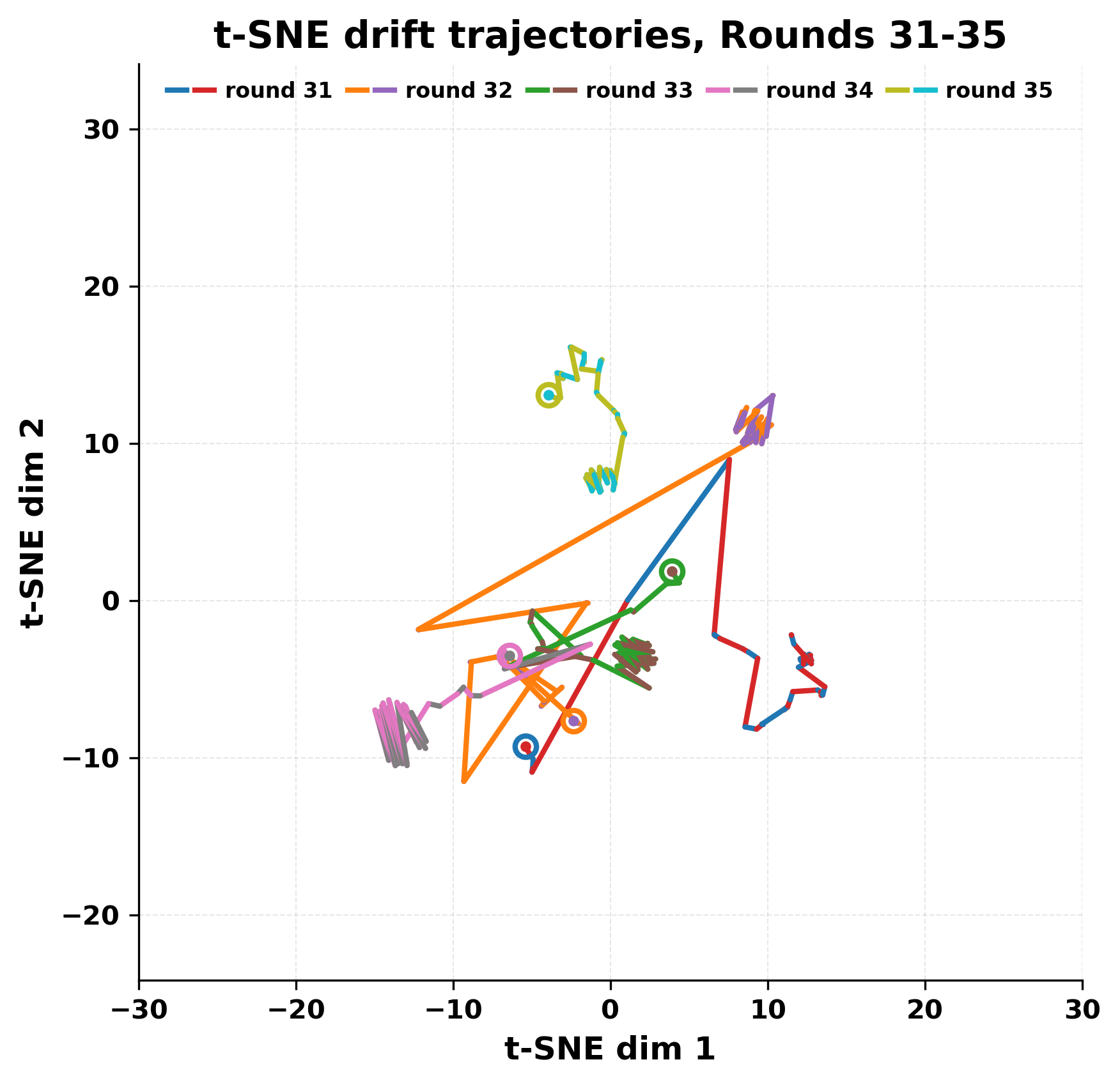}
    
    \includegraphics[width=\linewidth]{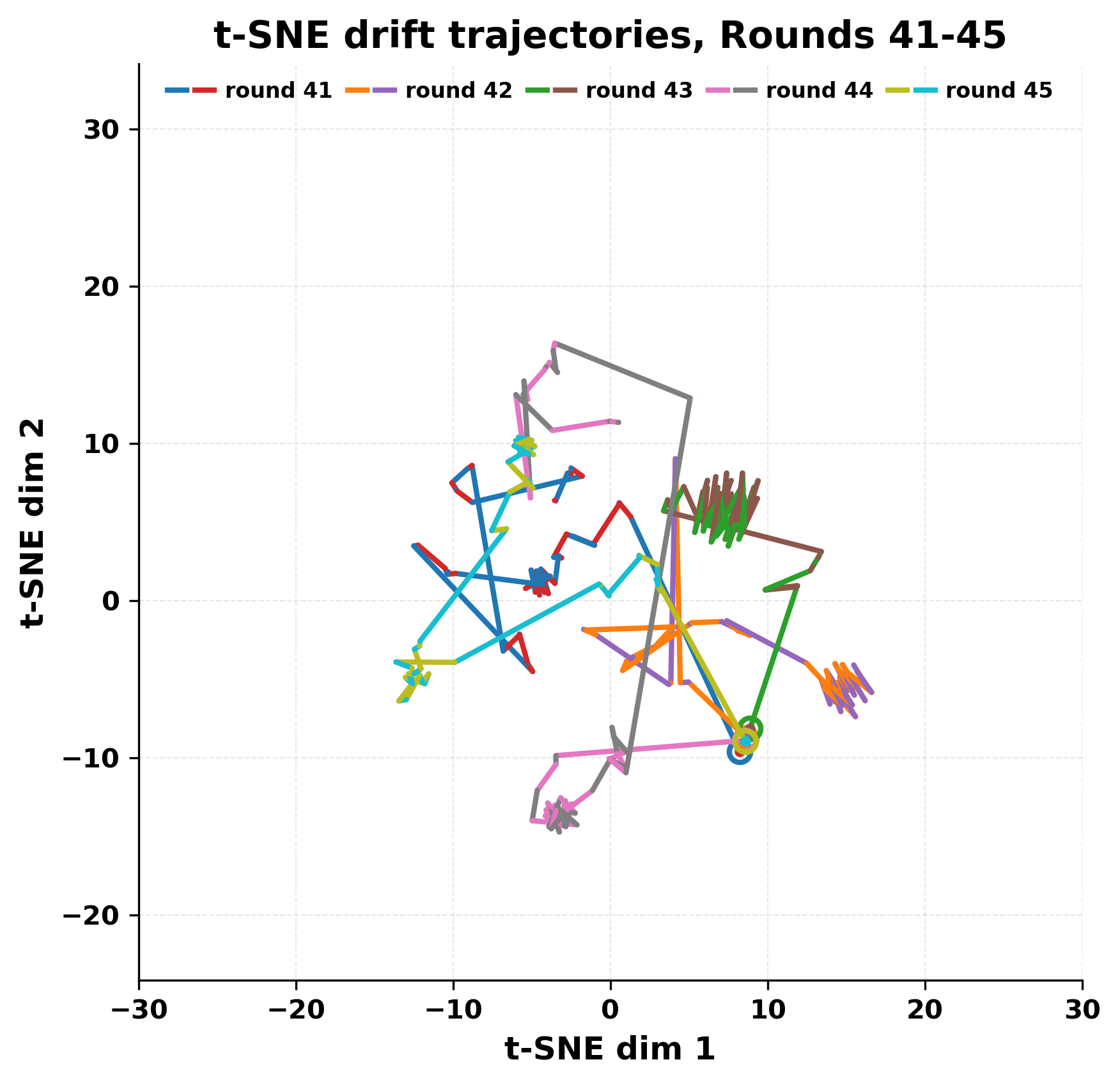}
\end{minipage}%
\hfill
\begin{minipage}[t]{0.50\textwidth}
    \centering
    \includegraphics[width=\linewidth]{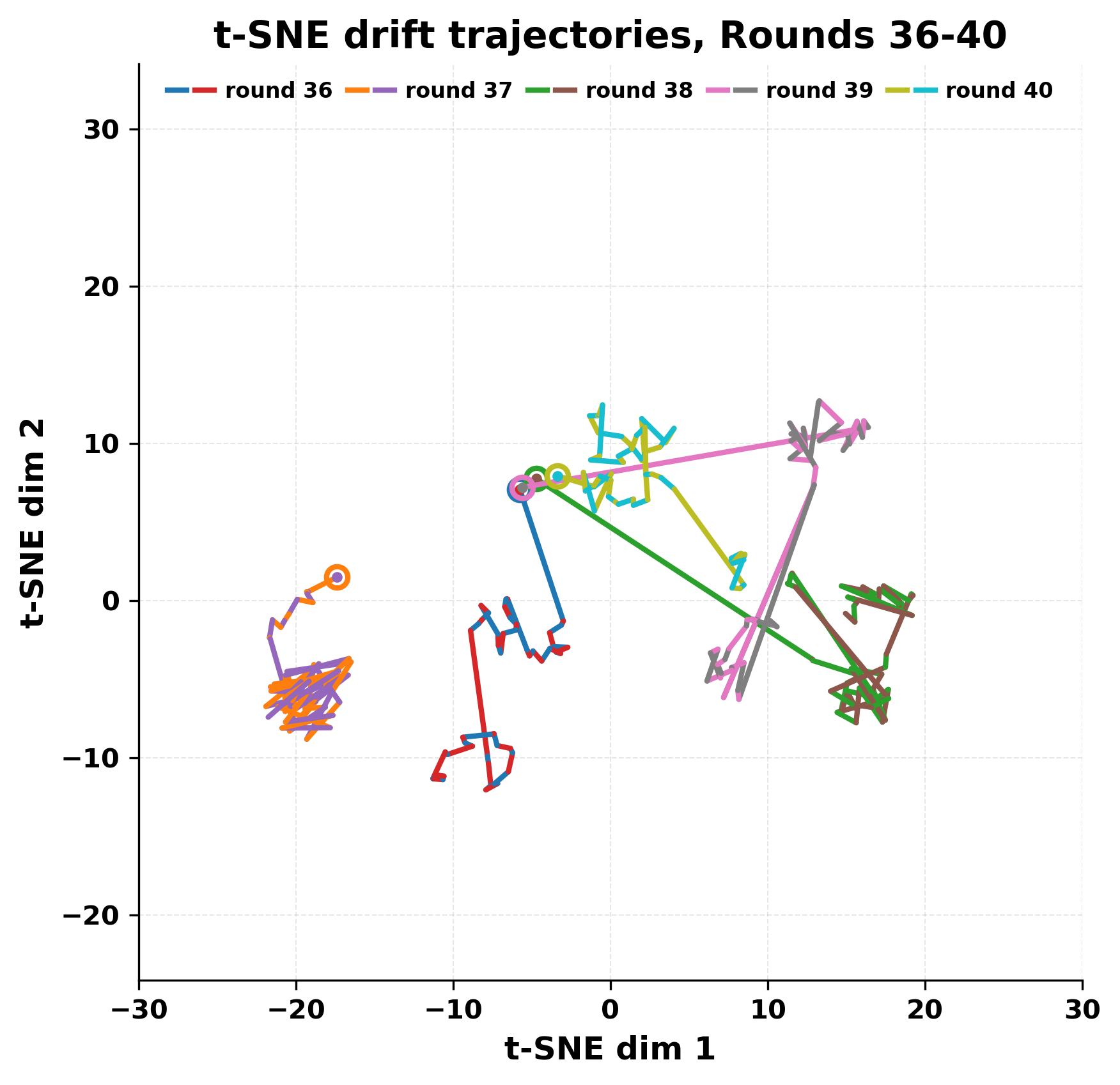}
    
    \includegraphics[width=\linewidth]{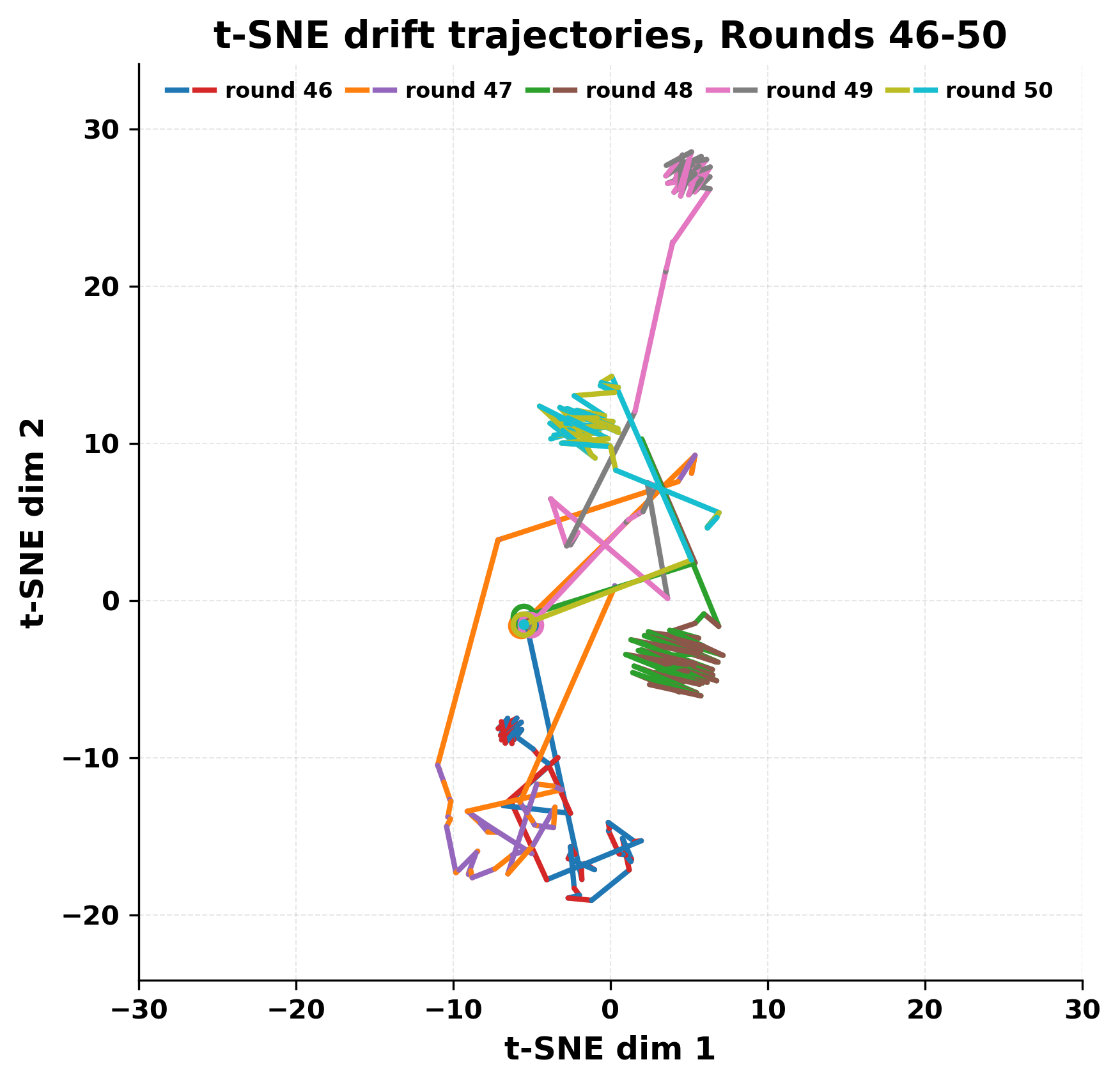}
\end{minipage}

\caption{t-SNE projection of sentence embeddings for Rounds 31–50, grouped in intervals of 5. Each point is a model output.}
\label{fig:tsne2}
\end{figure}

In Fig.\ref{fig:tsne1} and \ref{fig:tsne2}, the sequence of sentence embeddings are mapped to two-dimensional space (the t-SNE algorithm preserved local neighborhood relationships from the original space). We have shown the progression of the conversation by first creating the embedding of the generated sentences from the original sentence using the all-mpnet-base-v2 \cite{ashqar2023comparative} \cite{siino2024all} model. This model creates a vector of dimension $768$. Then we applied t-SNE to project these high-dimensional vectors in a lower-dimensional space (dimension=$2$) for visualization purposes. 

Starting from the origin sentence (marked as circle), we used the alternate coloring scheme as described earlier. The first color corresponds to Model A, Mistral, and the second color corresponds to Model B, LLaMA. It is evident from these figures that the convergence phenomena occurs frequently irrespective of data source of the seed sentence.

\subsubsection{Collapse Detection via Thresholding: }
To detect convergence without manual inspection, a rule-based method was applied on the per-step distances. The method checks for sequences of steps where the distance between consecutive outputs drops below a fixed cutoff. If the distance between outputs stays low for a fixed number of steps (window size=3) in a row, the conversation is marked as converged. The respective cutoff values for different metrics and the number of required consecutive steps are selected by comparing with manually labeled runs.

Cosine distance in embedding space provided the clearest signal. In most collapsed conversations, this value drops sharply and stays low. Similar trends appear in Jaccard and BLEU distances. Coherence scores sometimes fluctuate before flattening near the point of collapse.

\begin{figure}[htbp]
\centering

\begin{minipage}[t]{0.48\textwidth}
    \centering  
    \includegraphics[width=\linewidth]{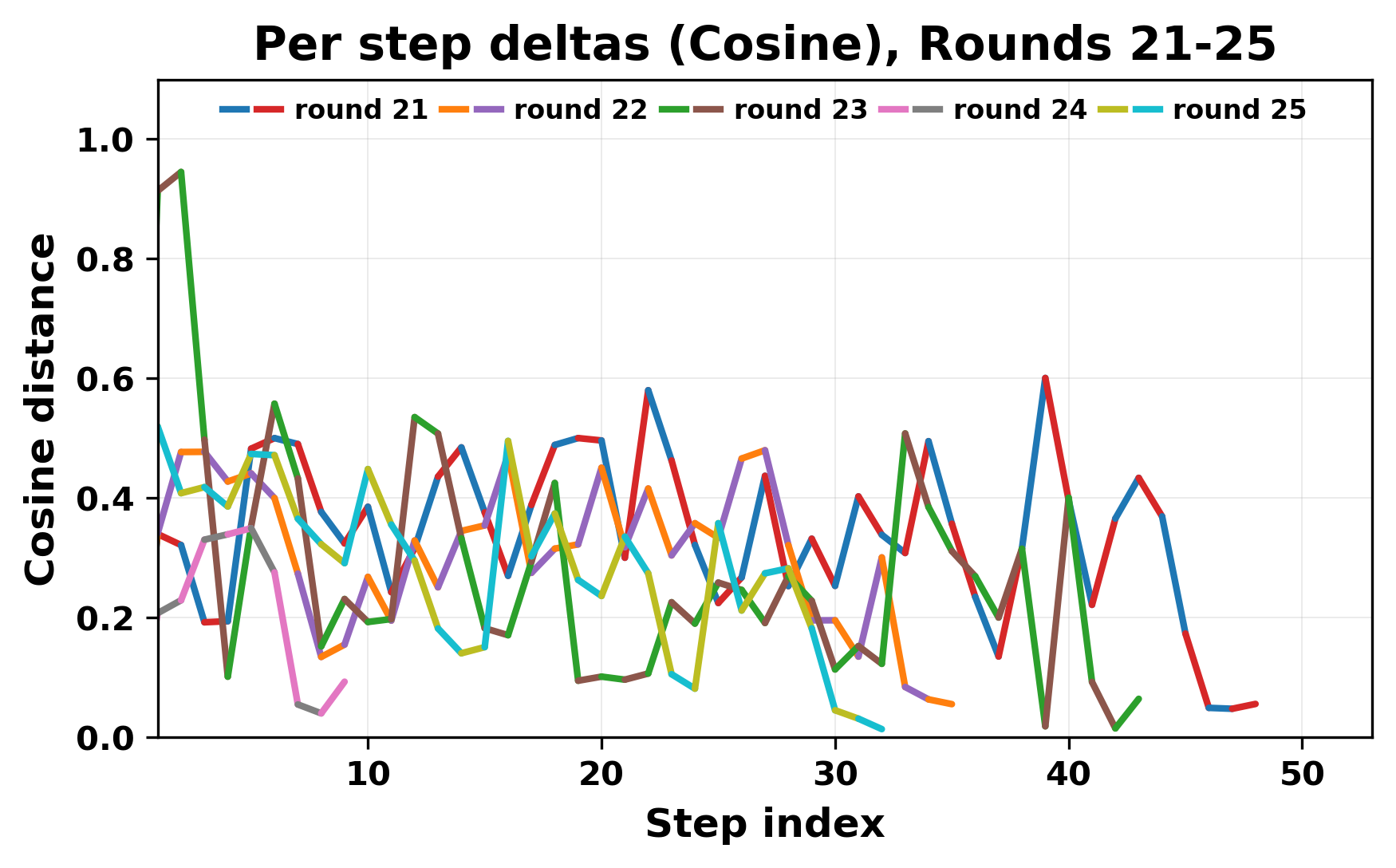}

\end{minipage}%
\hfill
\begin{minipage}[t]{0.48\textwidth}
    \centering   
    \includegraphics[width=\linewidth]{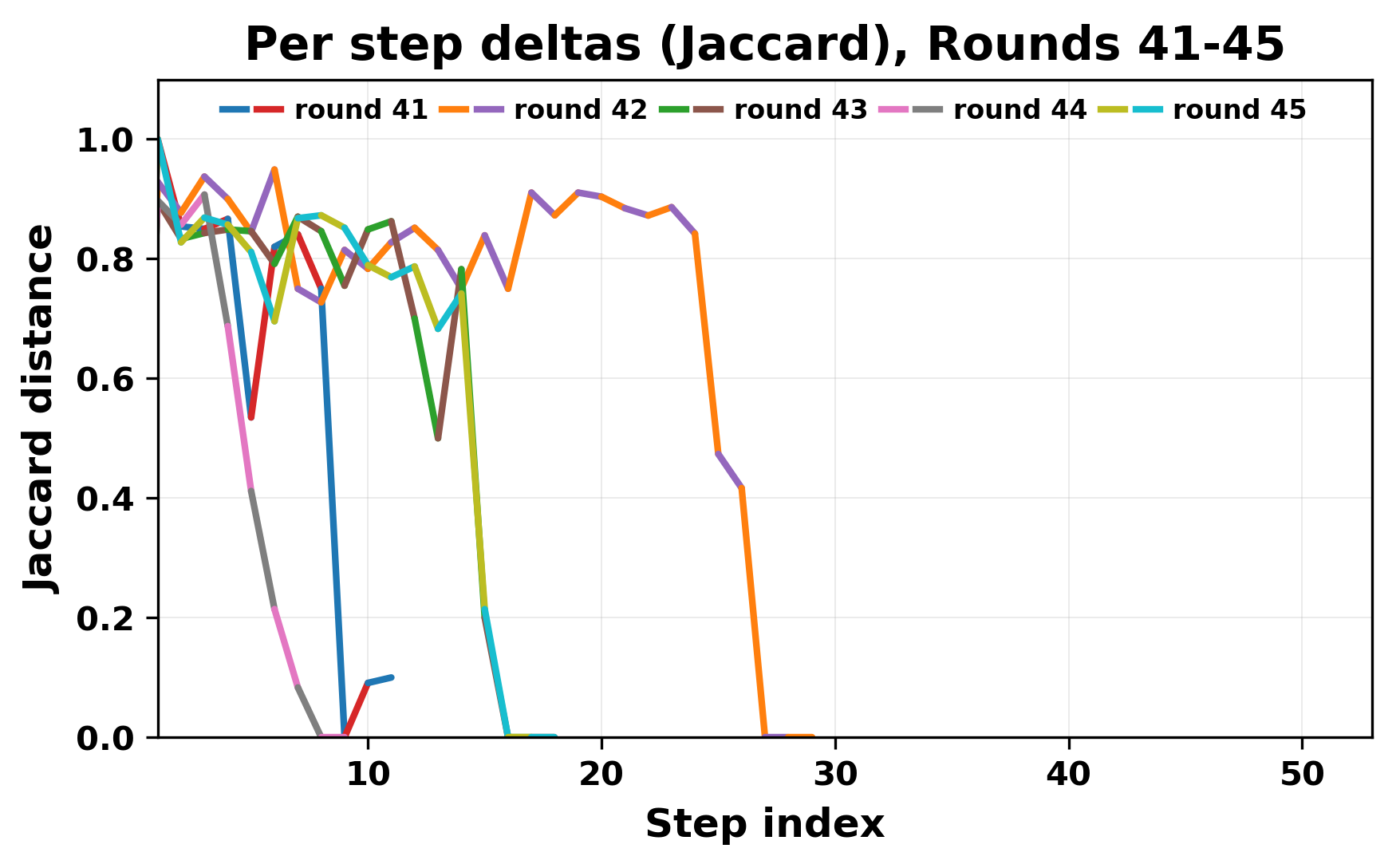}
    
\end{minipage}

\caption{After The application of Thresholding - Left: Metric is Cosine Distance (source: Prompt generated); Right: Metric is Jaccard Distance (source: Novels)}
\label{fig:cosine_jaccard}
\end{figure}

\begin{figure}[htbp]
\centering

\begin{minipage}[t]{0.48\textwidth}
    \centering
    \includegraphics[width=\linewidth]{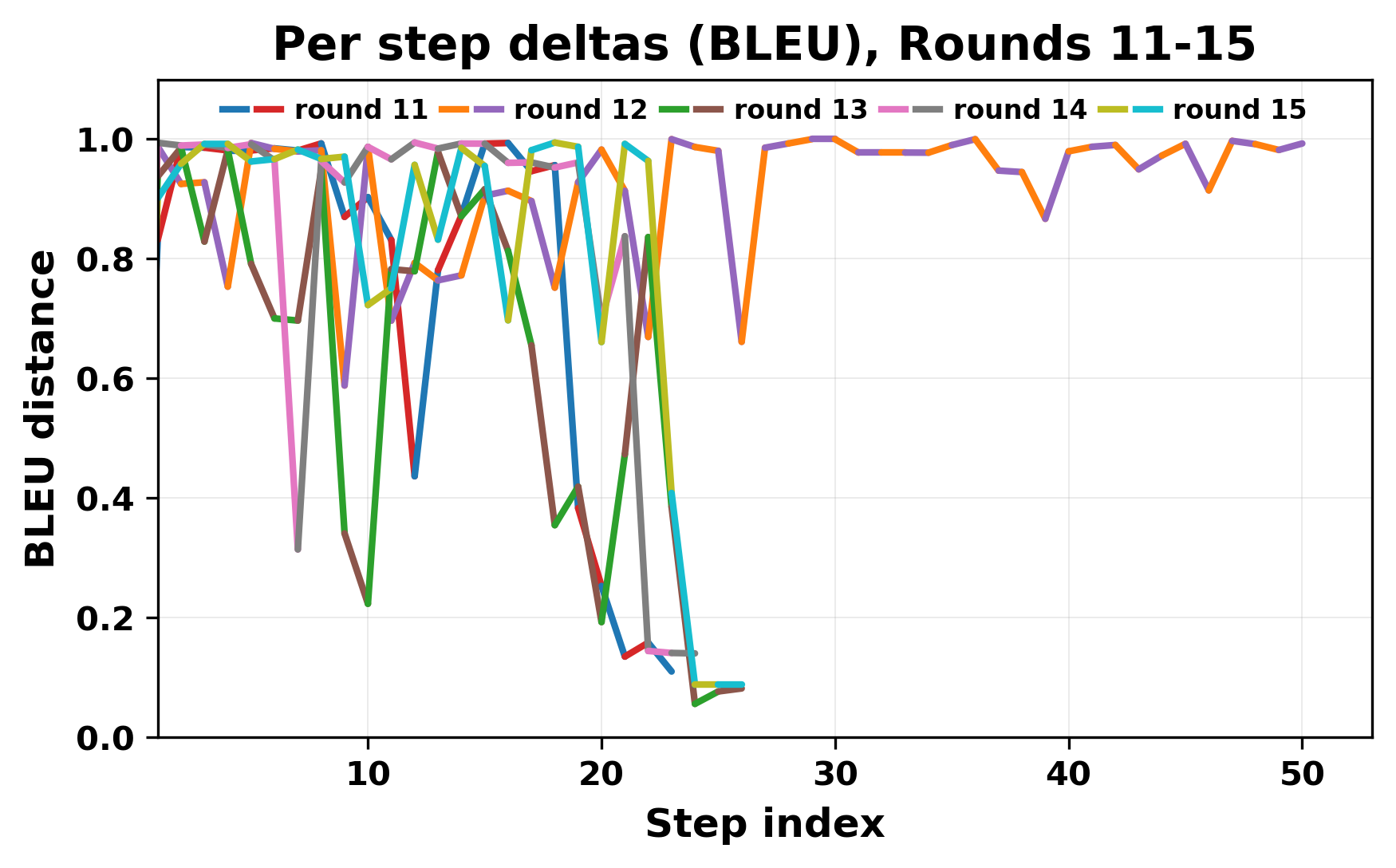}  
\end{minipage}%
\hfill
\begin{minipage}[t]{0.48\textwidth}
    \centering
    \includegraphics[width=\linewidth]{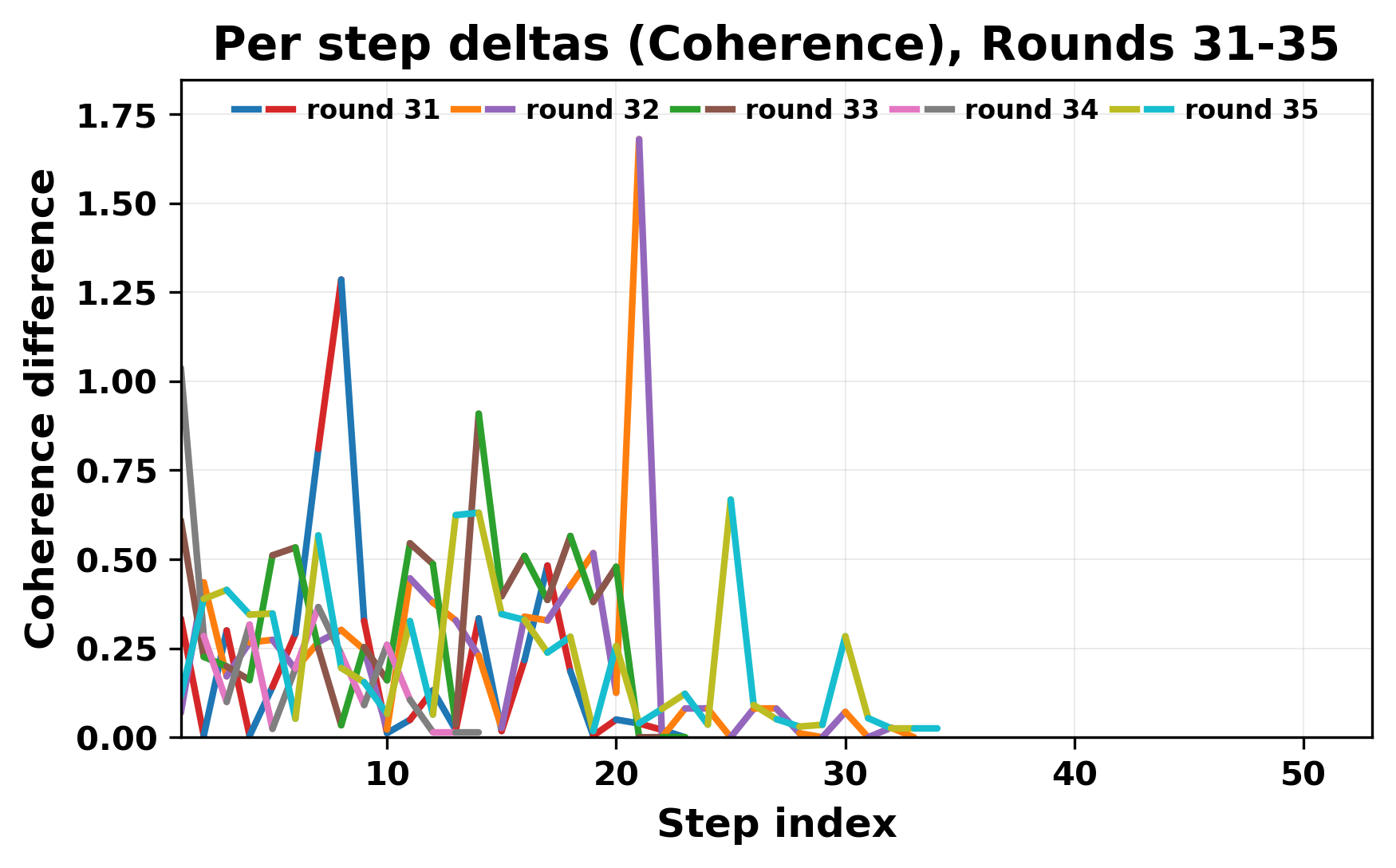}    
\end{minipage}

\caption{After the application of Thresholding -Left: Metric is BLEU score based distance (source: Wikipedia); Right: Metric is Coherence, (source: Scientific papers) }
\label{fig:bleu_coherence}
\end{figure}

We present a few example of this thresholding scheme in Fig. \ref{fig:cosine_jaccard} and \ref{fig:bleu_coherence}. We see that irrespective of the source, convergence occurs frequently. 

\FloatBarrier
\section{Conclusion}
This study examined how two large language models respond to each other over multiple steps using a variety of seed sentences. Many of the resulting conversations showed that after some steps the conversation becomes repetitive. Once repetition starts, both models continue to generate similar outputs as their inputs. This occurs without prompts, system instructions, or shared memory. 

Several new unexplored directions emerge from these findings. One is the use of external intervention to help a conversation escape low-diversity regions in language space. Another direction is model fine-tuning. A design that includes penalties for repetition or measures of novelty could change how models behave during such conversations. Token limits may also affect the behavior. We set the maximum output length to 50 tokens. This limit led to incomplete sentences. Despite that, the following model was able to continue the incomplete sentence. Testing longer outputs may change the pattern of repetition. Model size is another factor. This setup used two models with similar capacities. 

\section{Acknowledgment}
This research was partially supported by the NASA Established Program to Stimulate Competitive Research (EPSCoR), Grant No. 80NSSC22M0027 by the NASA West Virginia EPSCoR Committee.

\noindent Part of the computational requirements for this work were supported by resources at \textit{Delta at NCSA} and \textit{Anvil at Purdue University} through \textbf{ACCESS allocation Grant number: CIS250080} from the \textit{Advanced Cyberinfrastructure Coordination Ecosystem: Services and Support (ACCESS)} program, which is supported by U.S. National Science Foundation grants \#2138259, \#2138286, \#2138307, \#2137603, and \#2138296.

\bibliographystyle{splncs04}
\bibliography{references}

\end{document}